\documentclass[10pt,twocolumn,letterpaper]{article}

\usepackage{cvpr,lmodern,times,epsfig,graphicx,amsmath,amssymb}
\usepackage{xcolor,tabulary,xfrac,enumitem}
\usepackage[british,english,american]{babel}
\usepackage[pagebackref=true,breaklinks=true,letterpaper=true,colorlinks,
  citecolor=citecolor,bookmarks=false]{hyperref}
\usepackage[caption=false]{subfig}
\definecolor{citecolor}{RGB}{34,139,34}
\def\cvprPaperID{}\cvprfinalcopy

\renewcommand{\paragraph}[1]{\vspace{1.5mm}\noindent\textbf{#1}}
\newcommand{\eqnsm}[3]{\vspace{#1}\begin{equation}\label{eq:#2}#3\vspace{#1}\end{equation}\ignorespaces}
\newcommand{\tablestyle}[2]{\setlength{\tabcolsep}{#1}\renewcommand{\arraystretch}{#2}\centering\footnotesize}
\newlength\savewidth\newcommand\shline{\noalign{\global\savewidth\arrayrulewidth
  \global\arrayrulewidth 1pt}\hline\noalign{\global\arrayrulewidth\savewidth}}
\newcommand{\includegraphicstriml}[3]{\includegraphics[trim=#1 0 0 0, clip, #2]{#3}}

\newcommand{\app}{\raise.17ex\hbox{$\scriptstyle\sim$}}
\newcommand{\round}[1]{\lfloor#1\rceil}

\newcommand{\origerr}[1]{\color{gray}{[#1]}}
\newcommand{\mypm}[1]{\color{gray}{\tiny{$\pm$#1}}}

\def\x{{\times}}
\def\c{{\cdot}}
\newcommand{\g}{\color{gray}}
\newcolumntype{m}{>{\scshape}l}

\newcommand{\dsname}[1]{\texttt{\small #1}\xspace}
\newcommand{\dsnamebold}[1]{{\fontseries{b}\selectfont\texttt{#1}}\xspace}
\newcommand{\anynet}{\dsname{AnyNet}}
\newcommand{\anynetx}{\dsname{AnyNetX}}
\newcommand{\anynetxv}[1]{\dsname{AnyNetX\textsubscript{#1}}}
\newcommand{\regnet}{\dsname{RegNet}}
\newcommand{\regnetx}{\dsname{RegNetX}}
\newcommand{\regnety}{\dsname{RegNetY}}
\newcommand{\model}[1]{\textsc{\small#1}}

\begin{document}
\title{Designing Network Design Spaces}
\author{%
 Ilija Radosavovic \quad Raj Prateek Kosaraju \quad Ross Girshick \quad Kaiming He \quad Piotr Doll\'ar\\[2mm]
 Facebook AI Research (FAIR)}
\maketitle

\begin{abstract}
In this work, we present a new network design paradigm. Our goal is to help advance the understanding of network design and discover design principles that generalize across settings. Instead of focusing on designing individual network instances, we design network design spaces that parametrize populations of networks. The overall process is analogous to classic manual design of networks, but elevated to the design space level. Using our methodology we explore the structure aspect of network design and arrive at a low-dimensional design space consisting of simple, regular networks that we call \regnet. The core insight of the \regnet parametrization is surprisingly simple: widths and depths of good networks can be explained by a quantized linear function. We analyze the \regnet design space and arrive at interesting findings that do not match the current practice of network design. The \regnet design space provides simple and fast networks that work well across a wide range of flop regimes. Under comparable training settings and flops, the \regnet models outperform the popular EfficientNet models while being up to 5$\times$ faster on GPUs.
\end{abstract}\vspace{-3mm}

\section{Introduction}

Deep convolutional neural networks are the engine of visual recognition. Over the past several years better architectures have resulted in considerable progress in a wide range of visual recognition tasks. Examples include LeNet~\cite{LeCun1989}, AlexNet~\cite{Krizhevsky2012}, VGG~\cite{Simonyan2015}, and ResNet~\cite{He2016}. This body of work advanced both the effectiveness of neural networks as well as our \emph{understanding} of network design. In particular, the above sequence of works demonstrated the importance of convolution, network and data size, depth, and residuals, respectively. The outcome of these works is not just particular network instantiations, but also \emph{design principles} that can be generalized and applied to numerous settings.

While \emph{manual} network design has led to large advances, finding well-optimized networks manually can be challenging, especially as the number of design choices increases. A popular approach to address this limitation is neural architecture search (NAS). Given a \emph{fixed} search space of possible networks, NAS automatically finds a good model within the search space. Recently, NAS has received a lot of attention and shown excellent results~\cite{Zoph2017, Liu2019, Tan2019}.

\begin{figure}[t]\centering\vspace{-3mm}
\includegraphics[width=1\linewidth]{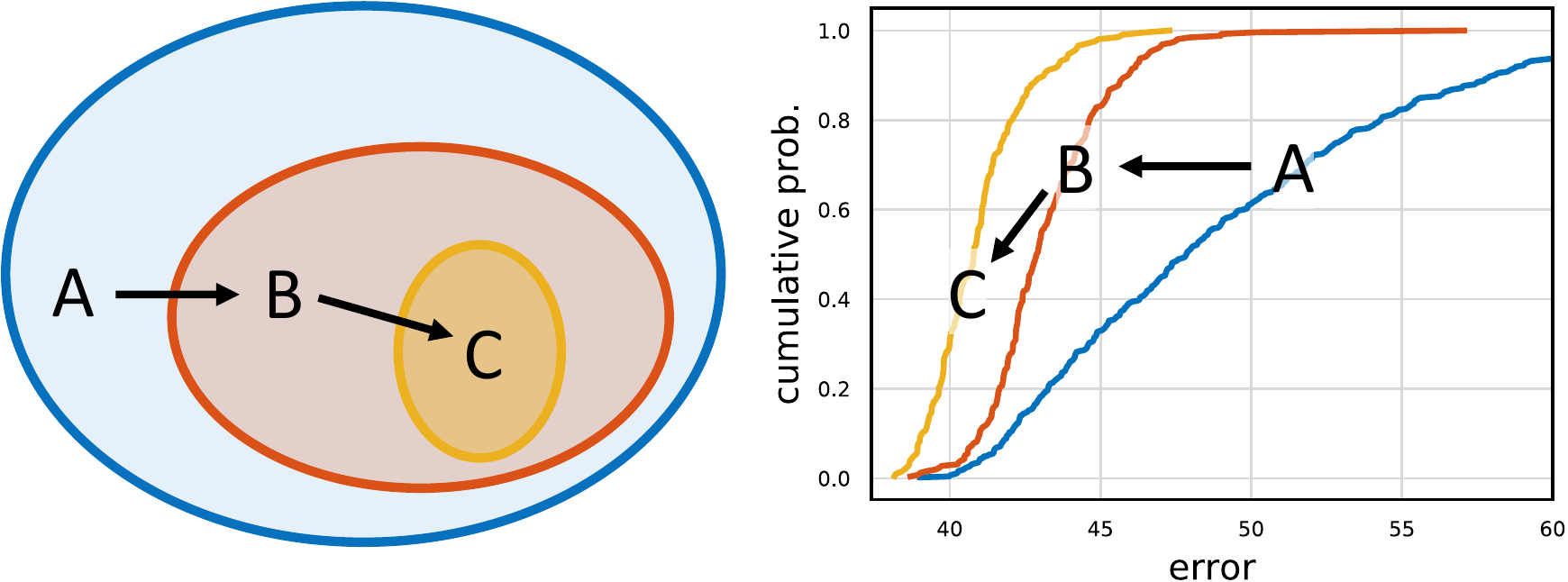}\vspace{-1mm}
\caption{\textbf{Design space design.} We propose to \emph{design} network design spaces, where a design space is a parametrized \emph{set} of possible model architectures. Design space design is akin to manual network design, but elevated to the \emph{population} level. In each step of our process the input is an initial design space and the output is a refined design space of simpler or better models. Following~\cite{Radosavovic2019}, we characterize the quality of a design space by sampling models and inspecting their \emph{error distribution}. For example, in the figure above we start with an initial design space \dsname{A} and apply two refinement steps to yield design spaces \dsname{B} then \dsname{C}. In this case \dsname{C} $\subseteq$ \dsname{B} $\subseteq$\dsname{A} (left), and the error distributions are strictly improving from \dsname{A} to \dsname{B} to \dsname{C} (right). The hope is that \emph{design principles} that apply to model populations are more likely to be robust and generalize.}
\label{fig:teaser}\vspace{-3mm}
\end{figure}

Despite the effectiveness of NAS, the paradigm has limitations. The outcome of the search is a single network \emph{instance} tuned to a specific setting (\eg, hardware platform). This is sufficient in some cases; however, it does not enable discovery of \emph{network design principles} that deepen our understanding and allow us to generalize to new settings. In particular, our aim is to find simple models that are easy to understand, build upon, and generalize.

In this work, we present a new network design paradigm that combines the advantages of manual design and NAS. Instead of focusing on designing individual network instances, we \emph{design design spaces} that parametrize populations of networks.\footnote{We use the term \emph{design space} following~\cite{Radosavovic2019}, rather than search space, to emphasize that we are not searching for network instances within the space. Instead, we are designing the space itself.} Like in manual design, we aim for interpretability and to discover general \emph{design principles} that describe networks that are simple, work well, and generalize across settings. Like in NAS, we aim to take advantage of semi-automated procedures to help achieve these goals.

The general strategy we adopt is to progressively design simplified versions of an initial, relatively unconstrained, design space while maintaining or improving its quality (Figure~\ref{fig:teaser}). The overall process is analogous to manual design, elevated to the population level and guided via distribution estimates of network design spaces~\cite{Radosavovic2019}.

As a testbed for this paradigm, our focus is on exploring network \emph{structure} (\eg, width, depth, groups, \etc) assuming standard model families including VGG~\cite{Simonyan2015}, ResNet~\cite{He2016}, and ResNeXt~\cite{Xie2017}. We start with a relatively unconstrained design space we call \anynet (\eg, widths and depths vary freely across stages) and apply our human-in-the-loop methodology to arrive at a low-dimensional design space consisting of simple ``regular'' networks, that we call \regnet. The core of the \regnet design space is simple: stage widths and depths are determined by a \emph{quantized linear function}. Compared to \anynet, the \regnet design space has simpler models, is easier to interpret, and has a higher concentration of good models.

We design the \regnet design space in a low-compute, low-epoch regime using a single network block type on ImageNet~\cite{Deng2009}. We then show that the \regnet design space generalizes to larger compute regimes, schedule lengths, and network block types. Furthermore, an important property of the design space design is that it is more interpretable and can lead to insights that we can learn from. We analyze the \regnet design space and arrive at interesting findings that do not match the current practice of network design. For example, we find that the depth of the best models is stable across compute regimes ($\app20$ blocks) and that the best models do not use either a bottleneck or inverted bottleneck.

We compare top \model{RegNet} models to existing networks in various settings. First, \model{RegNet} models are surprisingly effective in the mobile regime. We hope that these simple models can serve as strong baselines for future work. Next, \model{RegNet} models lead to considerable improvements over standard \model{ResNe(X)t}~\cite{He2016, Xie2017} models in all metrics. We highlight the improvements for fixed activations, which is of high practical interest as the number of activations can strongly influence the runtime on accelerators such as GPUs. Next, we compare to the state-of-the-art \model{EfficientNet}~\cite{Tan2019} models across compute regimes. Under \emph{comparable training settings} and flops, \model{RegNet} models outperform \model{EfficientNet} models while being up to \emph{5$\x$ faster} on GPUs. We further test generalization on ImageNetV2~\cite{Recht2019}.

We note that network structure is arguably the simplest form of a design space design one can consider. Focusing on designing richer design spaces (\eg, including operators) may lead to better networks. Nevertheless, the structure will likely remain a core component of such design spaces.

In order to facilitate future research we will release all code and pretrained models introduced in this work.\footnote{\url{https://github.com/facebookresearch/pycls}}

\section{Related Work}

\paragraph{Manual network design.} The introduction of AlexNet~\cite{Krizhevsky2012} catapulted network design into a thriving research area. In the following years, improved network designs were proposed; examples include VGG~\cite{Simonyan2015}, Inception~\cite{Szegedy2015, Szegedy2016a}, ResNet~\cite{He2016}, ResNeXt~\cite{Xie2017}, DenseNet~\cite{Huang2017}, and MobileNet~\cite{Howard2017, Sandler2018}. The design process behind these networks was largely manual and focussed on discovering new design choices that improve accuracy \eg, the use of deeper models or residuals. We likewise share the goal of discovering new design principles. In fact, our methodology is analogous to manual design but performed at the design space level.

\paragraph{Automated network design.} Recently, the network design process has shifted from a manual exploration to more automated network design, popularized by NAS\@. NAS has proven to be an effective tool for finding good models, \eg,~\cite{Zoph2018, Real2018, Liu2018, Pham2018, Liu2019, Tan2019}. The majority of work in NAS focuses on the search algorithm, \ie, efficiently finding the best network instances within a fixed, manually designed search space (which we call a \emph{design space}). Instead, our focus is on a paradigm for designing novel design spaces. The two are complementary: better design spaces can improve the efficiency of NAS search algorithms and also lead to existence of better models by enriching the design space.

\paragraph{Network scaling.} Both manual and semi-automated network design typically focus on finding best-performing network instances for a specific regime (\eg, number of flops comparable to ResNet-50). Since the result of this procedure is a single network instance, it is not clear how to adapt the instance to a different regime (\eg, fewer flops). A common practice is to apply network scaling rules, such as varying network depth~\cite{He2016}, width~\cite{Zagoruyko2016}, resolution~\cite{Howard2017}, or all three jointly~\cite{Tan2019}. Instead, our goal is to discover general design principles that hold across regimes and allow for efficient tuning for the optimal network in any target regime.

\paragraph{Comparing networks.} Given the vast number of possible network design spaces, it is essential to use a reliable comparison metric to guide our design process. Recently, the authors of~\cite{Radosavovic2019} proposed a methodology for comparing and analyzing populations of networks sampled from a design space. This distribution-level view is fully-aligned with our goal of finding general design principles. Thus, we adopt this methodology and demonstrate that it can serve as a useful tool for the design space design process.

\paragraph{Parameterization.} Our final quantized linear parameterization shares similarity with previous work, \eg how stage widths are set~\cite{Simonyan2015, He2015b, Zagoruyko2016, Huang2017, Howard2017}. However, there are two key differences. First, we provide an empirical study justifying the design choices we make. Second, we give insights into structural design choices that were not previously understood (\eg, how to set the number of blocks in each stages).

\section{Design Space Design}\label{sec:dds}

Our goal is to design better networks for visual recognition. Rather than designing or searching for a \emph{single} best model under specific settings, we study the behavior of \emph{populations} of models. We aim to discover general \emph{design principles} that can apply to and improve an entire model population. Such design principles can provide insights into network design and are more likely to generalize to new settings (unlike a single model tuned for a specific scenario).

We rely on the concept of network \emph{design spaces} introduced by Radosavovic \etal~\cite{Radosavovic2019}. A design space is a large, possibly infinite, population of model architectures. The core insight from~\cite{Radosavovic2019} is that we can sample models from a design space, giving rise to a model distribution, and turn to tools from classical statistics to analyze the design space. We note that this differs from architecture search, where the goal is to find the single best model from the space.

In this work, we propose to \emph{design} progressively simplified versions of an initial, unconstrained design space. We refer to this process as \emph{design space design}. Design space design is akin to sequential manual network design, but elevated to the population level. Specifically, in each step of our design process the input is an initial design space and the output is a refined design space, where the aim of each design step is to discover design principles that yield populations of simpler or better performing models.

We begin by describing the basic tools we use for design space design in \S\ref{sec:dds:tools}. Next, in \S\ref{sec:dds:anynet} we apply our methodology to a design space, called \anynet, that allows unconstrained network structures. In \S\ref{sec:dds:regnet}, after a sequence of design steps, we obtain a simplified design space consisting of only \emph{regular} network structures that we name \regnet. Finally, as our goal is not to design a design space for a single setting, but rather to discover general principles of network design that \emph{generalize} to new settings, in \S\ref{sec:dds:generalize} we test the generalization of the \regnet design space to new settings.

Relative to the \anynet design space, the \regnet design space is: (1) simplified both in terms of its dimension and type of network configurations it permits, (2) contains a higher concentration of top-performing models, and (3) is more amenable to analysis and interpretation.

\begin{figure}[t]\centering
\includegraphics[width=.32\linewidth]{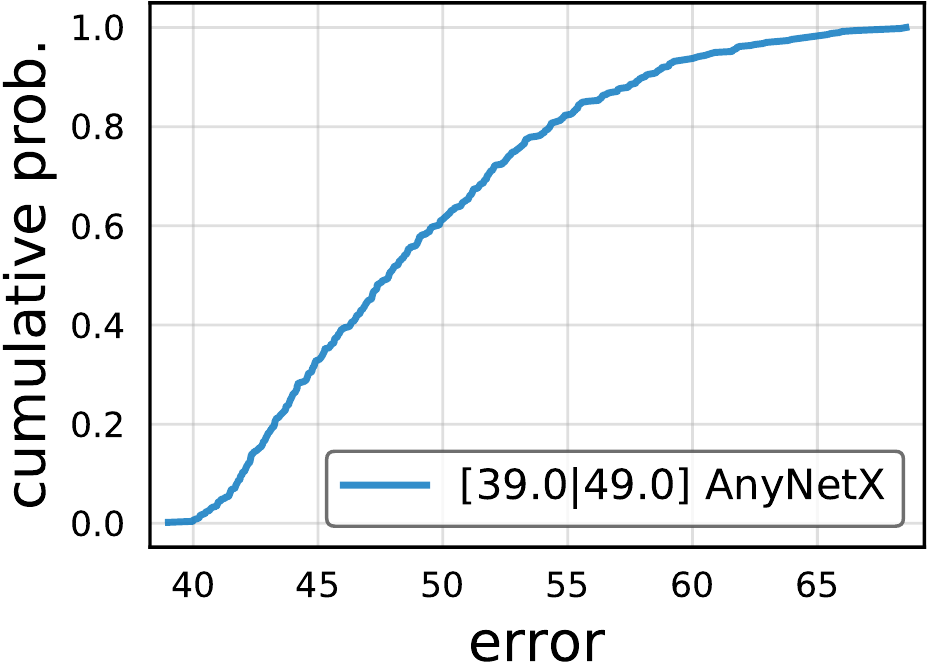}\hspace{.25mm}
\includegraphics[width=.32\linewidth]{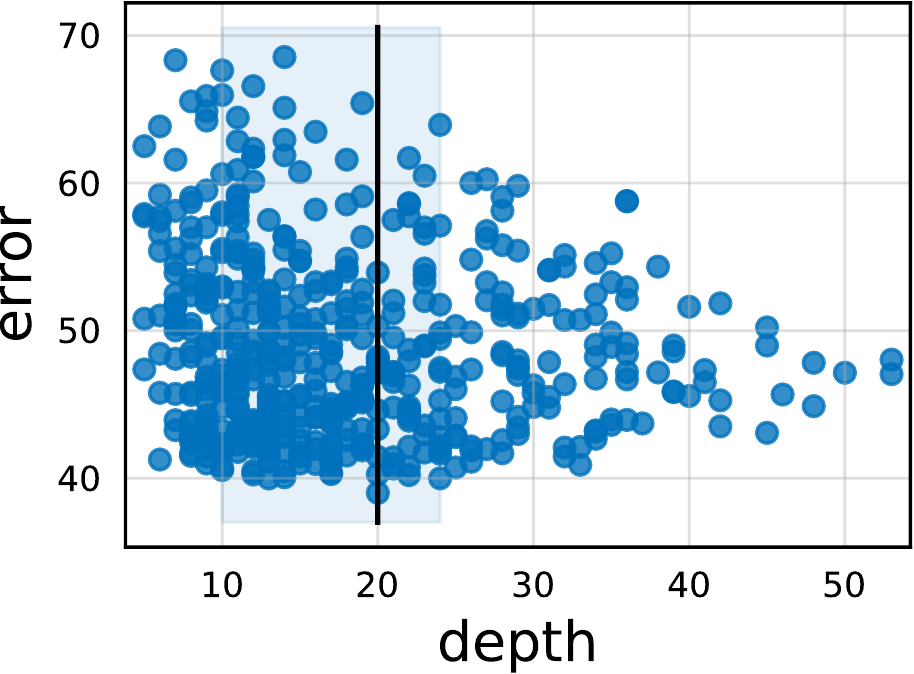}\hspace{.25mm}
\includegraphics[width=.32\linewidth]{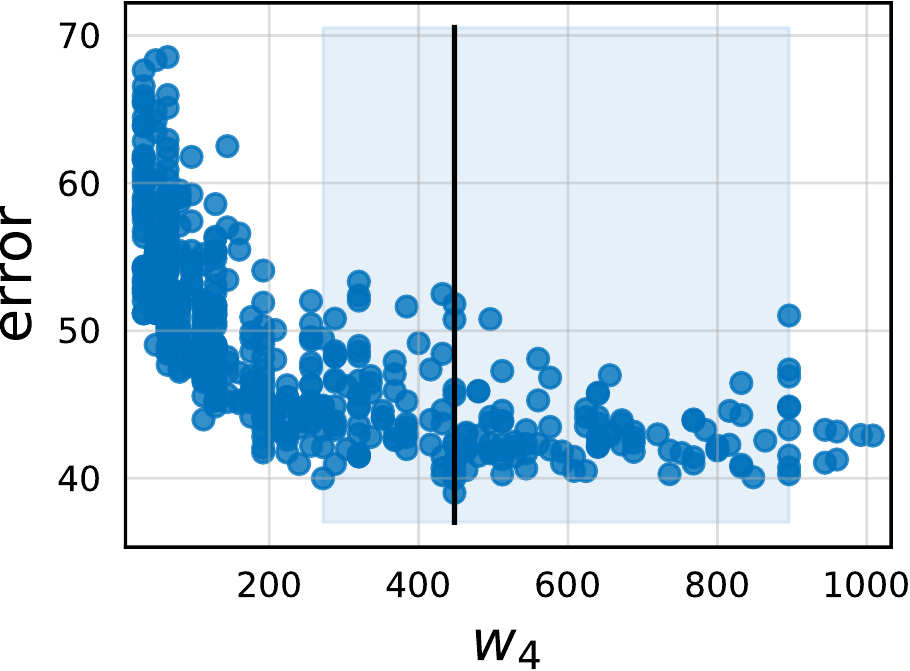}
\caption{Statistics of the \textbf{\anynetx design space} computed with $n=500$ sampled models. \textit{Left:} The error \emph{empirical distribution function} (EDF) serves as our foundational tool for visualizing the quality of the design space. In the legend we report the min error and mean error (which corresponds to the area under the curve). \textit{Middle:} Distribution of network depth $d$ (number of blocks) versus error. \textit{Right:} Distribution of block widths in the fourth stage ($w_4$) versus error. The blue shaded regions are ranges containing the best models with 95\% confidence (obtained using an empirical bootstrap), and the black vertical line the most likely best value.}
\label{fig:anynetx:a}\vspace{-3mm}
\end{figure}

\subsection{Tools for Design Space Design}\label{sec:dds:tools}

We begin with an overview of tools for design space design. To evaluate and compare design spaces, we use the tools introduced by Radosavovic \etal~\cite{Radosavovic2019}, who propose to quantify the quality of a design space by \emph{sampling} a set of models from that design space and characterizing the resulting model error \emph{distribution}. The key intuition behind this approach is that comparing distributions is more robust and informative than using search (manual or automated) and comparing the best found models from two design spaces.

To obtain a distribution of models, we sample and train $n$ models from a design space. For efficiency, we primarily do so in a \emph{low-compute}, \emph{low-epoch} training regime. In particular, in this section we use the 400 million flop\footnote{Following common practice, we use flops to mean multiply-adds. Moreover, we use MF and GF to denote $10^6$ and $10^9$ flops, respectively.} (400MF) regime and train each sampled model for 10 epochs on the ImageNet dataset~\cite{Deng2009}. We note that while we train many models, each training run is fast: training 100 models at 400MF for 10 epochs is roughly equivalent in flops to training a \emph{single} ResNet-50~\cite{He2016} model at 4GF for 100 epochs.

As in~\cite{Radosavovic2019}, our primary tool for analyzing design space quality is the error \emph{empirical distribution function} (EDF). The error EDF of $n$ models with errors $e_i$ is given by:
 \eqnsm{-1.5mm}{edf}{F(e)=\frac1n\sum_{i=1}^n \mathbf{1}[e_i < e].}
$F(e)$ gives the fraction of models with error less than $e$. We show the error EDF for $n=500$ sampled models from the \anynetx design space (described in \S\ref{sec:dds:anynet}) in Figure~\ref{fig:anynetx:a} (left).

Given a population of trained models, we can plot and analyze various network properties versus network error, see Figure~\ref{fig:anynetx:a} (middle) and (right) for two examples taken from the \anynetx design space. Such visualizations show 1D projections of a complex, high-dimensional space, and can help obtain insights into the design space. For these plots, we employ an \emph{empirical bootstrap}\footnote{Given $n$ pairs $(x_i, e_i)$ of model statistic $x_i$ (\eg depth) and corresponding error $e_i$, we compute the \emph{empirical bootstrap} by: (1) sampling with replacement 25\% of the pairs, (2) selecting the pair with min error in the sample, (3) repeating this $10^4$ times, and finally (4) computing the 95\% CI for the min $x$ value. The median gives the most likely best value.}~\cite{Efron1994} to estimate the likely range in which the best models fall.

To summarize: (1) we generate distributions of models obtained by sampling and training $n$ models from a design space, (2) we compute and plot error EDFs to summarize design space quality, (3) we visualize various properties of a design space and use an empirical bootstrap to gain insight, and (4) we use these insights to refine the design space.

\begin{figure}[t]\centering
\includegraphics[width=.85\linewidth]{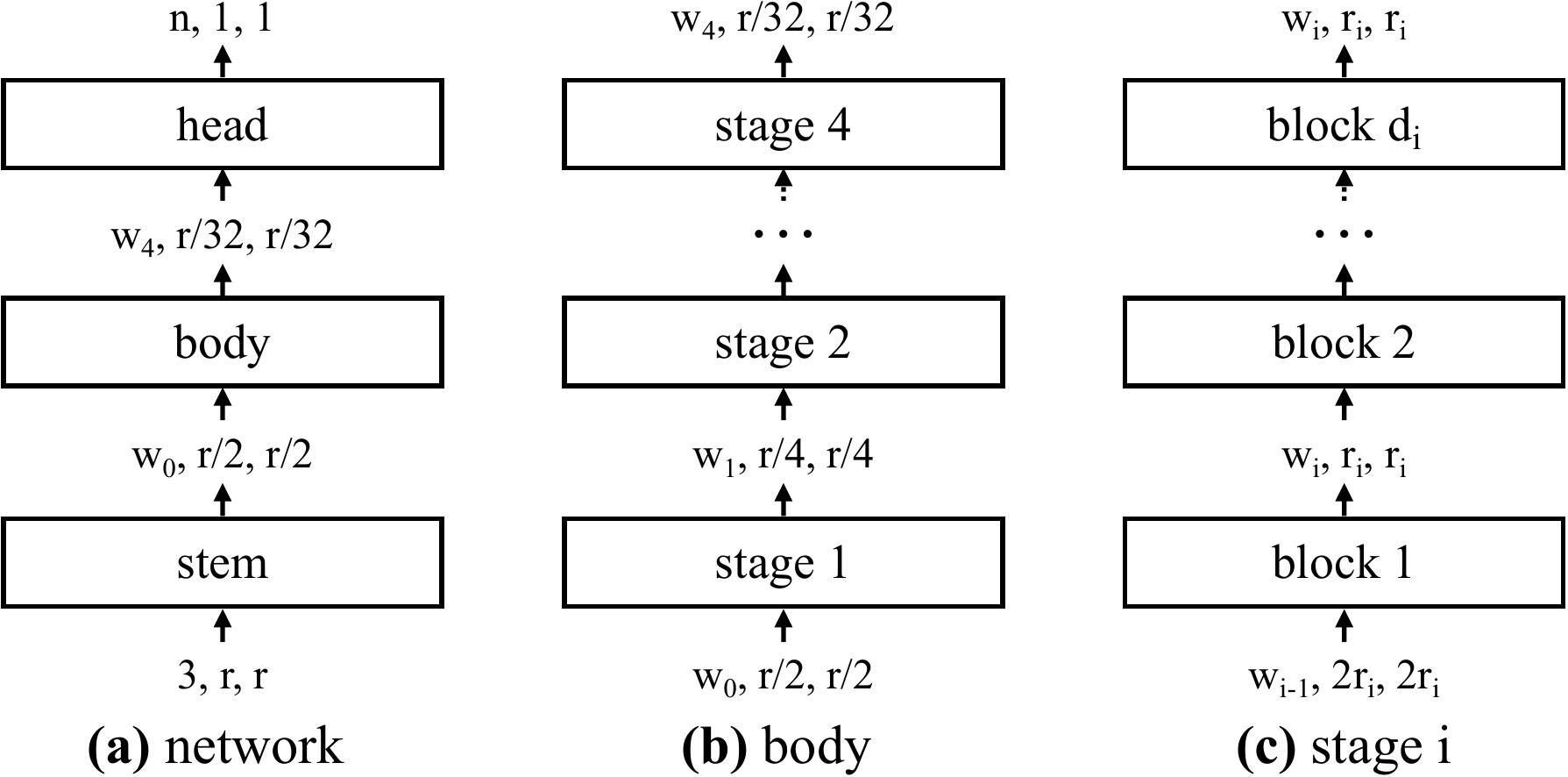}
\caption{General \textbf{network structure} for models in our design spaces. (a) Each network consists of a stem (stride-two $3\x3$ conv with $w_0=32$ output channels), followed by the network body that performs the bulk of the computation, and then a head (average pooling followed by a fully connected layer) that predicts $n$ output classes. (b) The network body is composed of a sequence of stages that operate at progressively reduced resolution $r_i$. (c) Each stage consists of a sequence of identical blocks, except the first block which uses stride-two conv. While the general structure is simple, the total number of possible network configurations is vast.}
\label{fig:structure:models}\vspace{-1mm}
\end{figure}

\subsection{The \dsnamebold{AnyNet} Design Space}\label{sec:dds:anynet}

We next introduce our initial \anynet design space. Our focus is on exploring the \emph{structure} of neural networks assuming standard, fixed network blocks (\eg, residual bottleneck blocks). In our terminology the structure of the network includes elements such as the number of blocks (\ie network depth), block widths (\ie number of channels), and other block parameters such as bottleneck ratios or group widths. The structure of the network determines the distribution of compute, parameters, and memory throughout the computational graph of the network and is key in determining its accuracy and efficiency.

The basic design of networks in our \anynet design space is straightforward. Given an input image, a network consists of a simple \emph{stem}, followed by the network \emph{body} that performs the bulk of the computation, and a final network \emph{head} that predicts the output classes, see Figure~\ref{fig:structure:models}a. We keep the stem and head fixed and as simple as possible, and instead focus on the structure of the network body that is central in determining network compute and accuracy.

The network body consists of 4 \emph{stages} operating at progressively reduced resolution, see Figure~\ref{fig:structure:models}b (we explore varying the number of stages in \S\ref{sec:dds:generalize}). Each stage consists of a sequence of identical blocks, see Figure~\ref{fig:structure:models}c. In total, for each stage $i$ the degrees of freedom include the number of blocks $d_i$, block width $w_i$, and any other block parameters. While the general structure is simple, the total number of possible networks in the \anynet design space is vast.

\begin{figure}[t]\centering\vspace{-2.5mm}
\includegraphics[width=.74\linewidth]{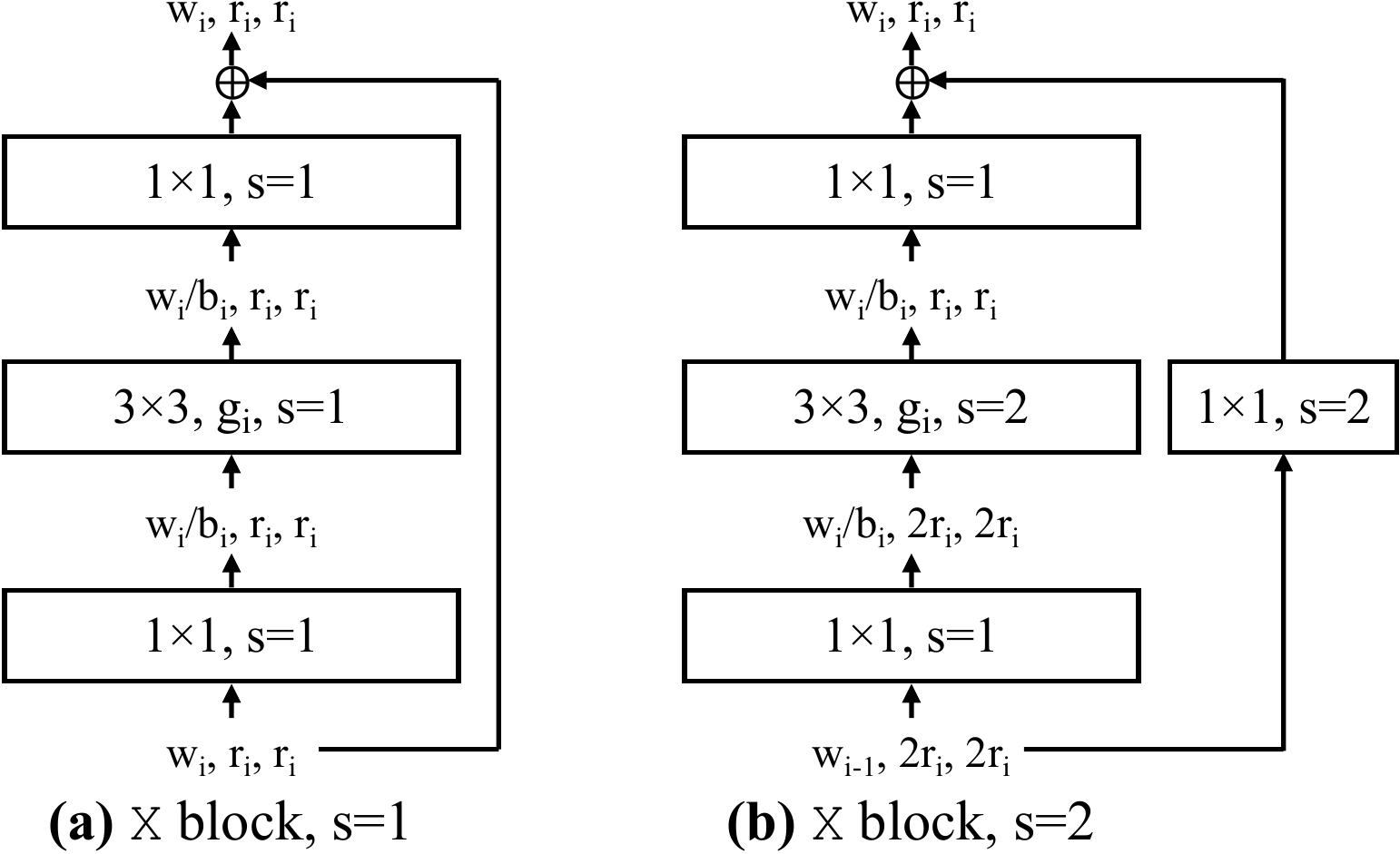}
\caption{The \textbf{\dsname{X} block} is based on the standard residual bottleneck block with group convolution~\cite{Xie2017}. (a) Each \dsname{X} block consists of a $1\x1$ conv, a $3\x3$ group conv, and a final $1\x1$ conv, where the $1\x1$ convs alter the channel width. BatchNorm~\cite{Ioffe2015} and ReLU follow each conv. The block has 3 parameters: the width $w_i$, bottleneck ratio $b_i$, and group width $g_i$. (b) The stride-two ($s=2$) version.}
\label{fig:structure:block:x}\vspace{-3.5mm}
\end{figure}

Most of our experiments use the standard residual bottlenecks block with group convolution~\cite{Xie2017}, shown in Figure~\ref{fig:structure:block:x}. We refer to this as the \dsname{X} block, and the \anynet design space built on it as \anynetx (we explore other blocks in \S\ref{sec:dds:generalize}). While the \dsname{X} block is quite rudimentary, we show it can be surprisingly effective when network structure is optimized.

The \anynetx design space has 16 degrees of freedom as each network consists of 4 stages and each stage $i$ has 4 parameters: the number of blocks $d_i$, block width $w_i$, bottleneck ratio $b_i$, and group width $g_i$. We fix the input resolution $r=224$ unless otherwise noted. To obtain valid models, we perform log-uniform sampling of $d_i\le16$, $w_i\le1024$ and divisible by 8, $b_i\in\{1, 2, 4\}$, and $g_i\in\{1, 2, \ldots, 32\}$ (we test these ranges later). We repeat the sampling until we obtain $n=500$ models in our target complexity regime (360MF to 400MF), and train each model for 10 epochs.\footnote{Our training setup in \S\ref{sec:dds} exactly follows~\cite{Radosavovic2019}. We use SGD with momentum of 0.9, mini-batch size of 128 on 1 GPU, and a half-period cosine schedule with initial learning rate of 0.05 and weight decay of $5\c10^{-5}$. Ten epochs are usually sufficient to give robust \emph{population} statistics.} Basic statistics for \anynetx were shown in Figure~\ref{fig:anynetx:a}.

There are $(16\c128\c3\c6)^4 \approx 10^{18}$ possible model configurations in the \anynetx design space. Rather than \emph{searching} for the single best model out of these $\app10^{18}$ configurations, we explore whether there are general \emph{design principles} that can help us understand and refine this design space. To do so, we apply our approach of designing design spaces. In each step of this approach, our aims are:
\begin{enumerate}[topsep=2.5pt, itemsep=-0.8ex]
\item to simplify the structure of the design space,
\item to improve the interpretability of the design space,
\item to improve or maintain the design space quality,
\item to maintain model diversity in the design space.
\end{enumerate}
We now apply this approach to the \anynetx design space.

\begin{figure}[t]\centering
\includegraphics[height=21.5mm]{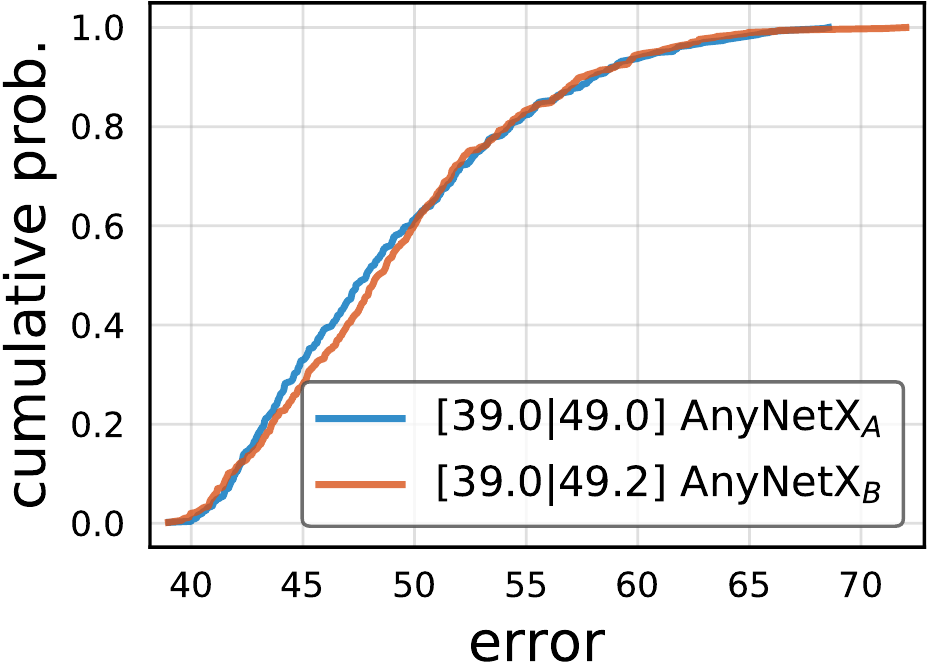}
\includegraphicstriml{40}{height=21.5mm}{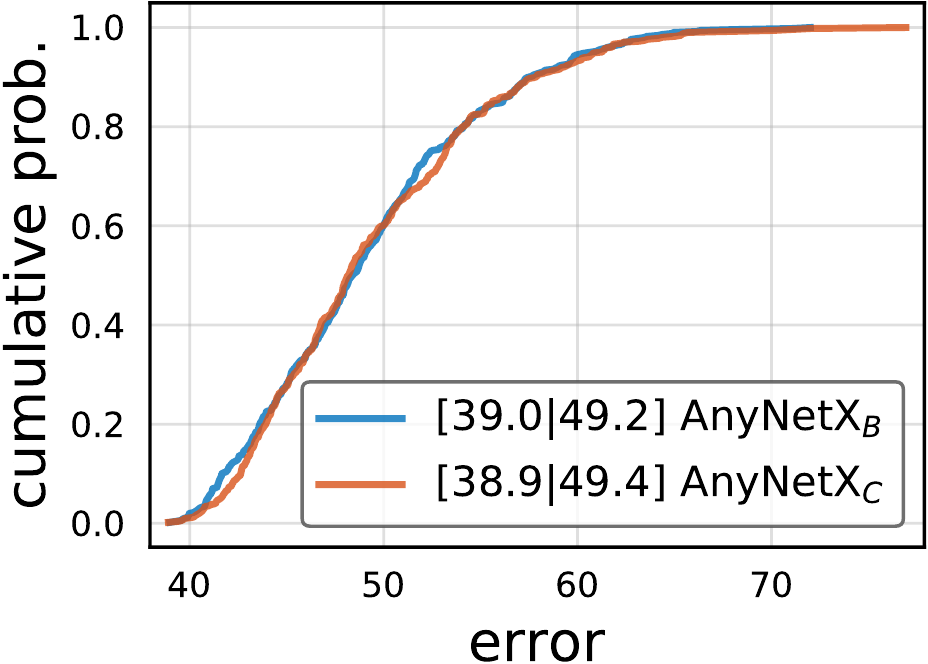}
\includegraphics[height=21.5mm]{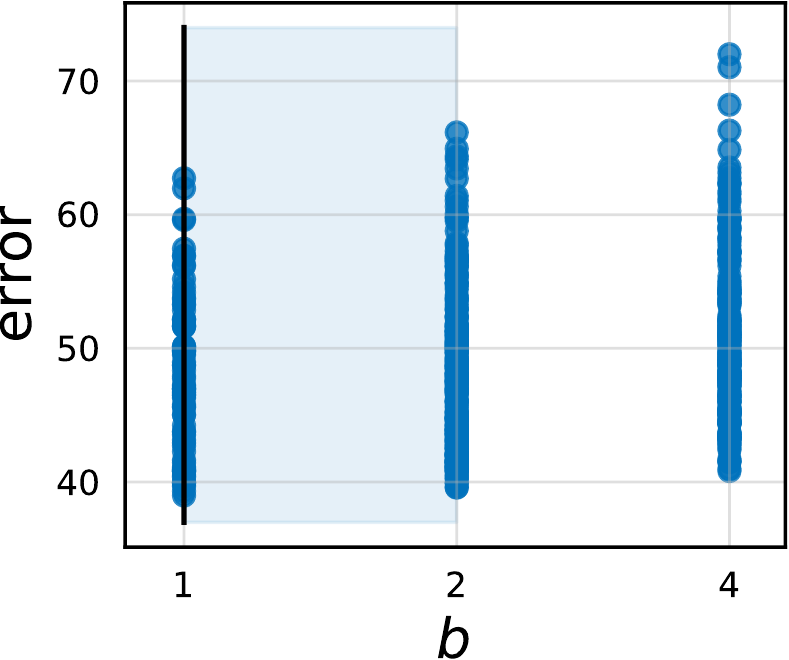}
\caption{\textbf{\anynetxv{B}} (left) and \textbf{\anynetxv{C}} (middle) introduce a \emph{shared} bottleneck ratio $b_i=b$ and shared group width $g_i=g$, respectively. This simplifies the design spaces while resulting in virtually no change in the error EDFs. Moreover, \anynetxv{B} and \anynetxv{C} are more amendable to analysis. Applying an empirical bootstrap to $b$ and $g$ we see trends emerge, \eg, with 95\% confidence $b\le2$ is best in this regime (right). No such trends are evident in the individual $b_i$ and $g_i$ in \anynetxv{A} (not shown).}
\label{fig:anynetx:bc}\vspace{-1mm}
\end{figure}

\paragraph{\anynetxv{A}.} For clarity, going forward we refer to the initial, unconstrained \anynetx design space as \anynetxv{A}.

\paragraph{\anynetxv{B}.} We first test a \emph{shared} bottleneck ratio $b_i=b$ for all stages $i$ for the \anynetxv{A} design space, and refer to the resulting design space as \anynetxv{B}. As before, we sample and train 500 models from \anynetxv{B} in the same settings. The EDFs of \anynetxv{A} and \anynetxv{B}, shown in Figure~\ref{fig:anynetx:bc} (left), are virtually identical both in the average and best case. This indicates no loss in accuracy when coupling the $b_i$. In addition to being simpler, the \anynetxv{B} is more amenable to analysis, see for example Figure~\ref{fig:anynetx:bc} (right).

\paragraph{\anynetxv{C}.} Our second refinement step closely follows the first. Starting with \anynetxv{B}, we additionally use a \emph{shared} group width $g_i=g$ for all stages to obtain \anynetxv{C}. As before, the EDFs are nearly unchanged, see Figure~\ref{fig:anynetx:bc} (middle). Overall, \anynetxv{C} has 6 fewer degrees of freedom than \anynetxv{A}, and reduces the design space size nearly four orders of magnitude. Interestingly, we find $g>1$ is best (not shown); we analyze this in more detail in \S\ref{sec:regnet}.

\begin{figure}[t]\centering
\includegraphics[width=\linewidth]{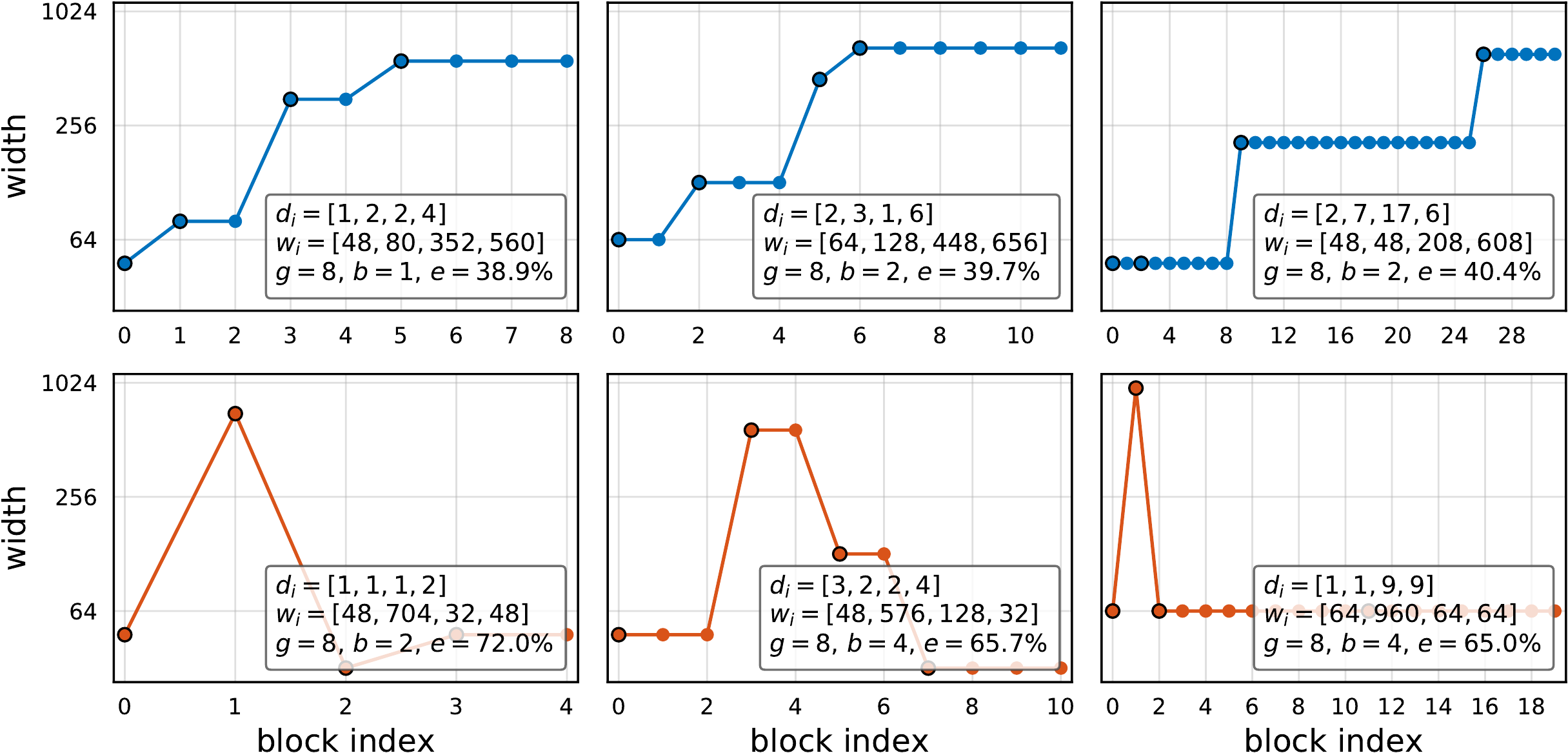}
\caption{Example good and bad \textbf{\anynetxv{C} networks}, shown in the top and bottom rows, respectively. For each network, we plot the width $w_j$ of every block $j$ up to the network depth $d$. These \emph{per-block} widths $w_j$ are computed from the \emph{per-stage} block depths $d_i$ and block widths $w_i$ (listed in the legends for reference).}
\label{fig:models:anynetx}\vspace{-3mm}
\end{figure}

\begin{figure}[t]\centering
\includegraphics[height=22mm]{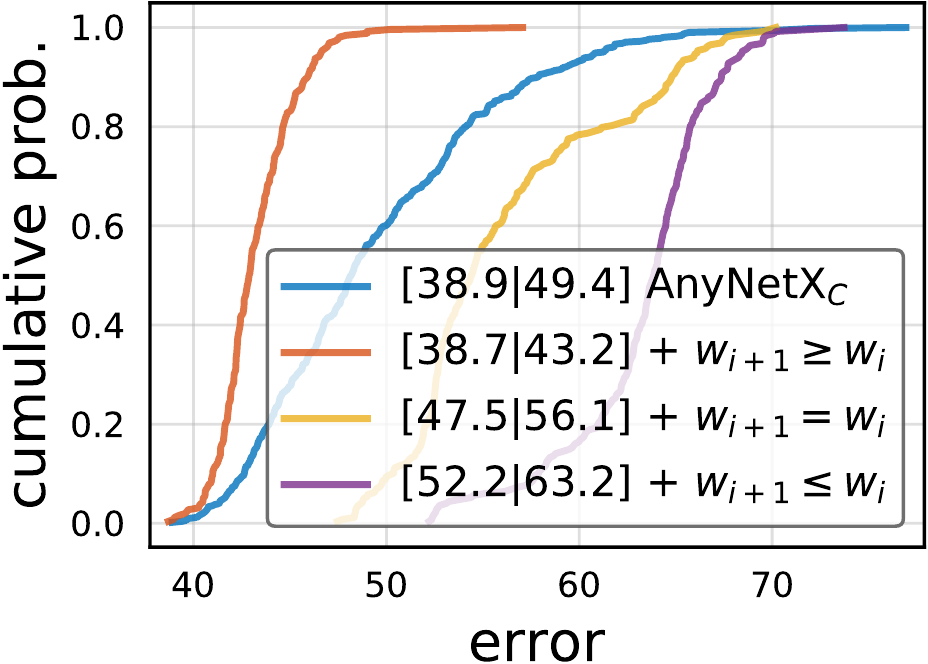}\hspace{2mm}
\includegraphicstriml{40}{height=22mm}{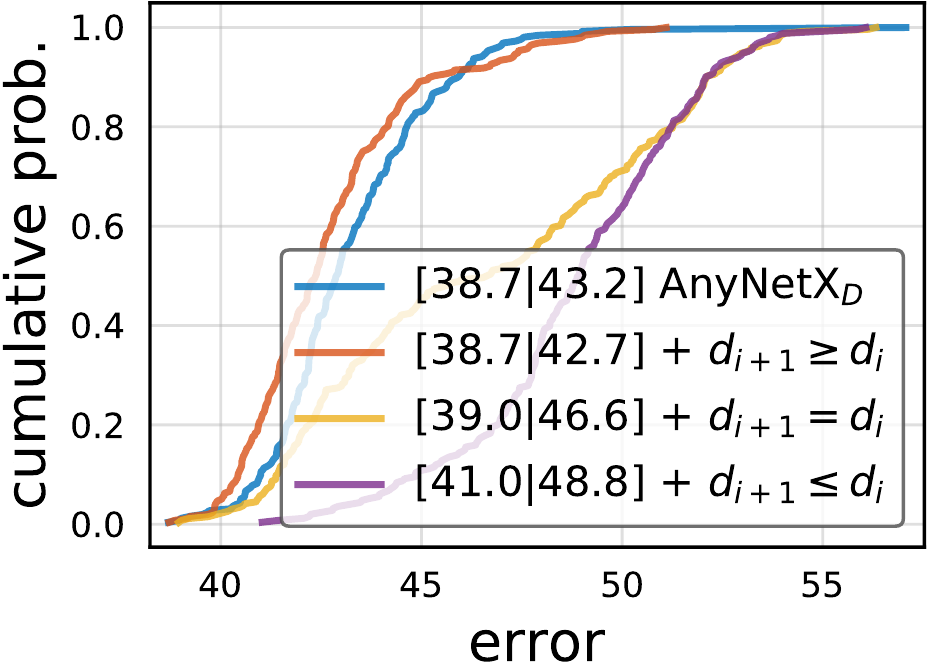}
\caption{\textbf{\anynetxv{D}} (left) and \textbf{\anynetxv{E}} (right). We show various constraints on the per stage widths $w_i$ and depths $d_i$. In both cases, having increasing $w_i$ and $d_i$ is beneficial, while using constant or decreasing values is much worse. Note that \anynetxv{D} = \anynetxv{C} + $w_{i+1} \ge w_i$, and \anynetxv{E} = \anynetxv{D} + $d_{i+1} \ge d_i$. We explore stronger constraints on $w_i$ and $d_i$ shortly.}
\label{fig:anynetx:de}
\end{figure}

\paragraph{\anynetxv{D}.} Next, we examine typical network structures of both good and bad networks from \anynetxv{C} in Figure~\ref{fig:models:anynetx}. A pattern emerges: good network have increasing widths. We test the design principle of $w_{i+1}\ge w_i$, and refer to the design space with this constraint as \anynetxv{D}. In Figure~\ref{fig:anynetx:de} (left) we see this improves the EDF substantially. We return to examining other options for controlling width shortly.

\paragraph{\anynetxv{E}.} Upon further inspection of many models (not shown), we observed another interesting trend. In addition to stage widths $w_i$ increasing with $i$, the stage depths $d_i$ likewise tend to increase for the best models, although not necessarily in the last stage. Nevertheless, we test a design space variant \anynetxv{E} with $d_{i+1}\ge d_i$ in Figure~\ref{fig:anynetx:de} (right), and see it also improves results. Finally, we note that the constraints on $w_i$ and $d_i$ each reduce the design space by $4!$, with a cumulative reduction of $O(10^7)$ from \anynetxv{A}.

\begin{figure}[t]\centering
\hspace{-2mm}
\includegraphics[height=21.3mm]{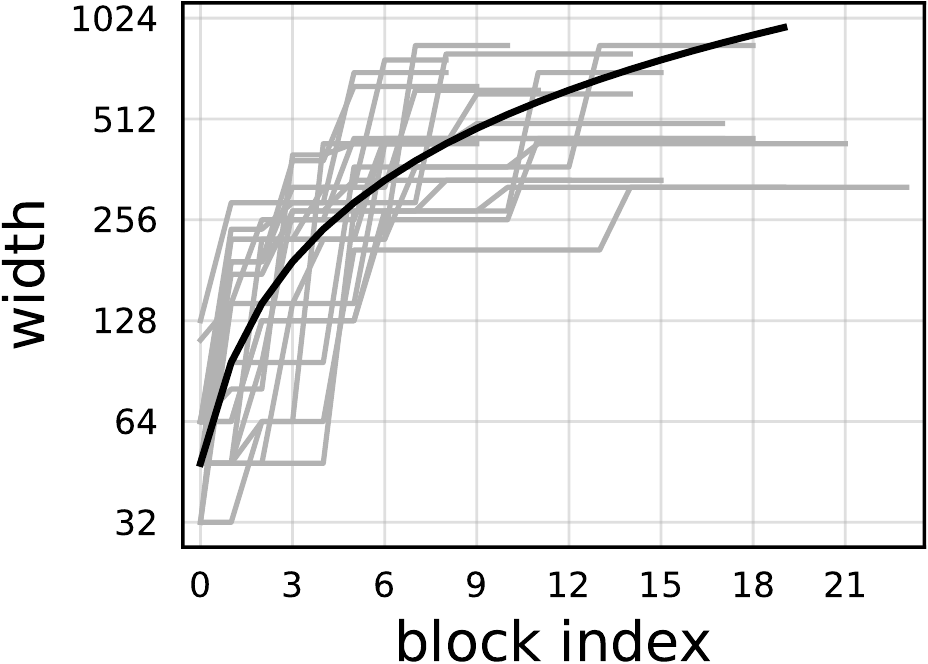}\hspace{.25mm}
\includegraphics[height=21.3mm]{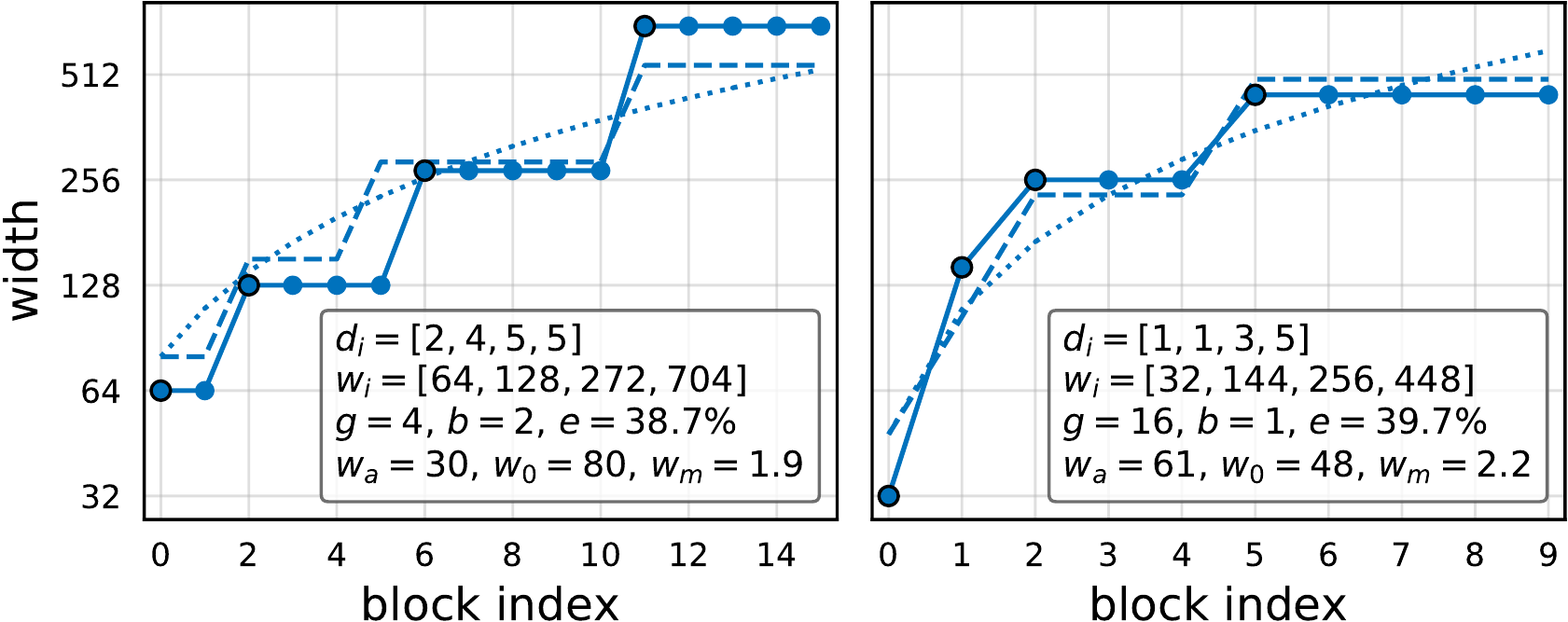}\\[1mm]
\includegraphics[width=\linewidth]{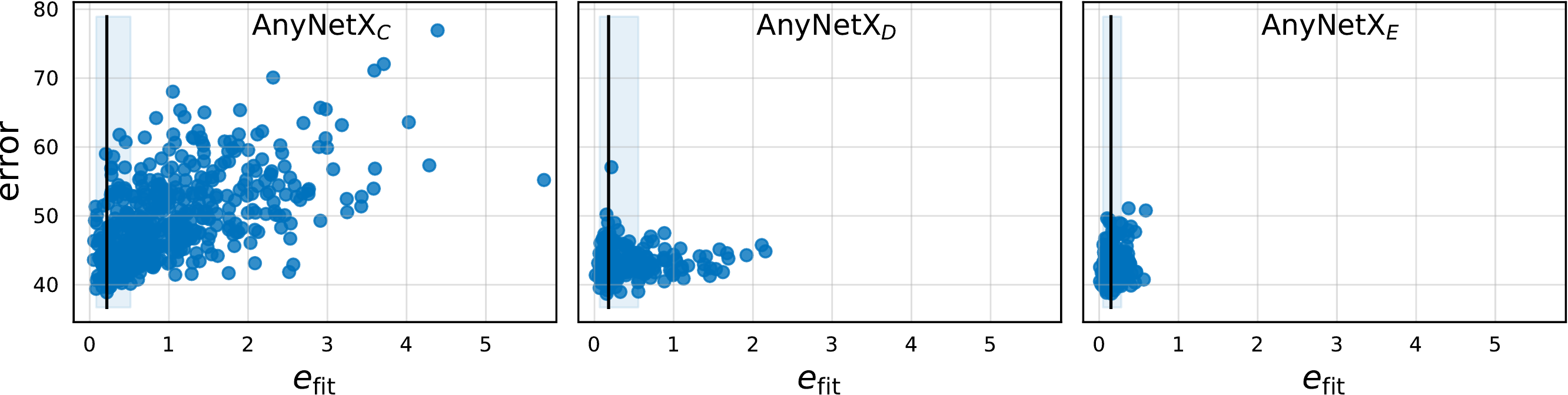}
\caption{\textbf{Linear fits}. Top networks from the \anynetx design space can be well modeled by a \emph{quantized linear parameterization}, and conversely, networks for which this parameterization has a higher fitting error $e_\textrm{fit}$ tend to perform poorly. See text for details.}
\label{fig:linear}\vspace{-3mm}
\end{figure}

\subsection{The \dsnamebold{RegNet} Design Space}\label{sec:dds:regnet}

To gain further insight into the model structure, we show the best 20 models from \anynetxv{E} in a single plot, see Figure~\ref{fig:linear} (top-left). For each model, we plot the per-block width $w_j$ of every block $j$ up to the network depth $d$ (we use $i$ and $j$ to index over stages and blocks, respectively). See Figure~\ref{fig:models:anynetx} for reference of our model visualization.

While there is significant variance in the individual models (gray curves), in the aggregate a pattern emerges. In particular, in the same plot we show the line $w_j=48 \cdot (j + 1)$ for $0 \le j \le 20$ (solid black curve, please note that the y-axis is logarithmic). Remarkably, this trivial \emph{linear fit} seems to explain the population trend of the growth of network widths for top models. Note, however, that this linear fit assigns a different width $w_j$ to each block, whereas individual models have quantized widths (piecewise constant functions).

To see if a similar pattern applies to \emph{individual} models, we need a strategy to \emph{quantize} a line to a piecewise constant function. Inspired by our observations from \anynetxv{D} and \anynetxv{E}, we propose the following approach. First, we introduce a \emph{linear parameterization} for block widths:
 \eqnsm{-2mm}{linear}{u_j = w_0 + w_a \cdot j\quad\textrm{for}\quad 0 \le j<d}
This parameterization has three parameters: depth $d$, initial width $w_0 > 0$, and slope $w_a > 0$, and generates a different block width $u_j$ for each block $j<d$. To quantize $u_j$, we introduce an additional parameter $w_m > 0$ that controls quantization as follows. First, given $u_j$ from Eqn.~(\ref{eq:linear}), we compute $s_j$ for each block j such that the following holds:
 \eqnsm{-2mm}{continuous}{u_j = w_0 \cdot w_m^{s_j}}
Then, to quantize $u_j$, we simply round $s_j$ (denoted by $\round{s_j}$) and compute quantized per-block widths $w_j$ via:
 \eqnsm{-2mm}{quantize}{w_j = w_0 \cdot w_m^{\round{s_j}}}
We can convert the per-block $w_j$ to our per-stage format by simply counting the number of blocks with constant width, that is, each stage $i$ has block width $w_i = w_0 \c w_m^i$ and number of blocks $d_i = \sum_j\mathbf{1}[\round{s_j}=i$]. When only considering four stage networks, we ignore the parameter combinations that give rise to a different number of stages.

We test this parameterization by fitting to models from $\anynetx$. In particular, given a model, we compute the fit by setting $d$ to the network depth and performing a grid search over $w_0$, $w_a$ and $w_m$ to minimize the mean log-ratio (denoted by $e_\textrm{fit}$) of predicted to observed per-block widths. Results for two top networks from \anynetxv{E} are shown in Figure~\ref{fig:linear} (top-right). The quantized linear fits (dashed curves) are good fits of these best models (solid curves).

\begin{figure}[t]\centering
\includegraphics[height=20mm]{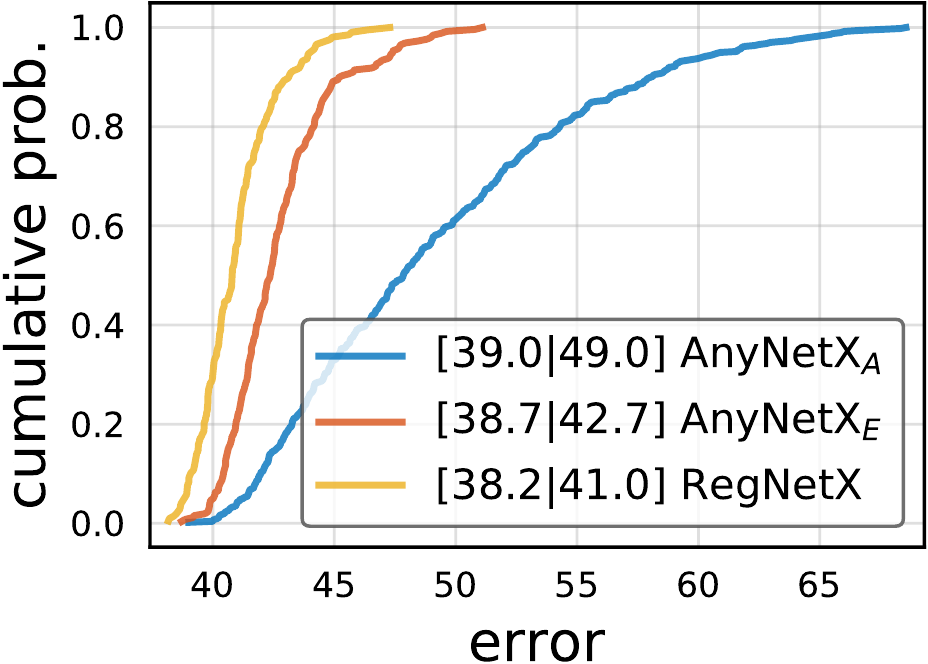}\hspace{1mm}
\includegraphicstriml{40}{height=20mm}{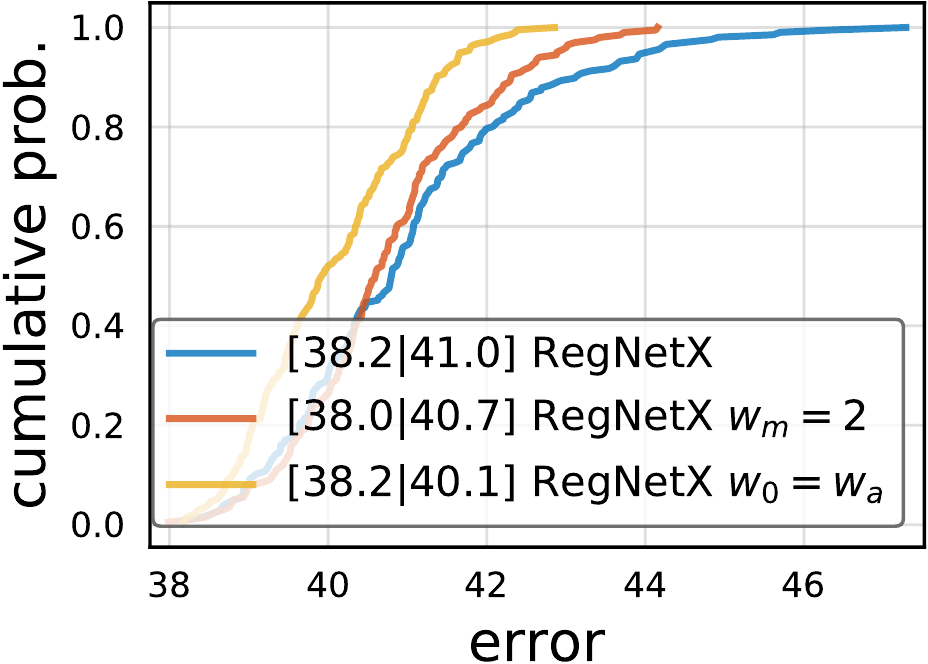}
\includegraphics[height=20mm]{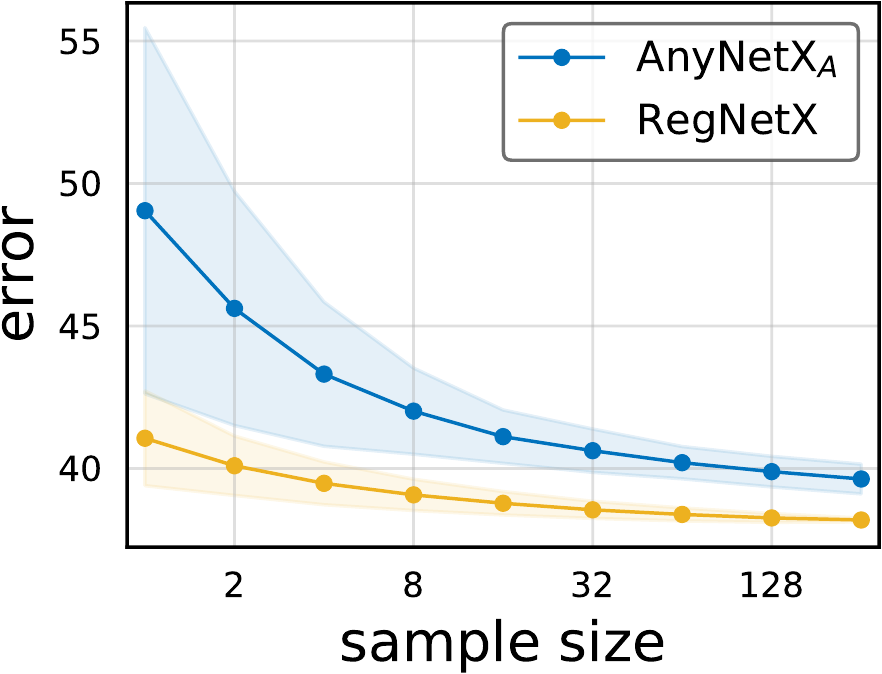}\
\caption{\textbf{\regnetx} design space. See text for details.}
\label{fig:regnetx}\vspace{-4mm}
\end{figure}

Next, we plot the fitting error $e_\textrm{fit}$ versus network error for every network in \anynetxv{C} through \anynetxv{E} in Figure~\ref{fig:linear} (bottom). First, we note that \emph{the best models in each design space all have good linear fits}. Indeed, an empirical bootstrap gives a narrow band of $e_\textrm{fit}$ near 0 that likely contains the best models in each design space. Second, we note that on average, $e_\textrm{fit}$ improves going from \anynetxv{C} to \anynetxv{E}, showing that the linear parametrization naturally enforces related constraints to $w_i$ and $d_i$ increasing.

To further test the linear parameterization, we design a design space that only contains models with such linear structure. In particular, we specify a network structure via 6 parameters: $d$, $w_0$, $w_a$, $w_m$ (and also $b$, $g$). Given these, we generate block widths and depths via Eqn.~(\ref{eq:linear})-(\ref{eq:quantize}). We refer to the resulting design space as \regnet, as it contains only simple, \emph{regular} models. We sample $d<64$, $w_0, w_a<256$, $1.5 \le w_m \le 3$ and $b$ and $g$ as before (ranges set based on $e_\textrm{fit}$ on \anynetxv{E}).

The error EDF of \regnetx is shown in Figure~\ref{fig:regnetx} (left). Models in \regnetx have better average error than \anynetx while maintaining the best models. In Figure~\ref{fig:regnetx} (middle) we test two further simplifications. First, using $w_m=2$ (doubling width between stages) slightly improves the EDF, but we note that using $w_m\ge2$ performs better (shown later). Second, we test setting $w_0=w_a$, further simplifying the linear parameterization to $u_j = w_a \cdot (j+1)$. Interestingly, this performs even better. However, to maintain the diversity of models, we do not impose either restriction. Finally, in Figure~\ref{fig:regnetx} (right) we show that random search efficiency is much higher for \regnetx; searching over just \app32 random models is likely to yield good models.

\begin{table}[t]\centering
\resizebox{.9\columnwidth}{!}{
\tablestyle{8pt}{1.05}
\begin{tabular}{@{}l|l|c|l|l@{}}
  & restriction & dim. & combinations & total \\\shline
 \anynetxv{A} & none                & 16 & $(16\c128\c3\c6)^4$        & \app$1.8\c10^{18}$\\
 \anynetxv{B} & + $b_{i+1} = b_i$   & 13 & $(16\c128\c6)^4\c3$        & \app$6.8\c10^{16}$\\
 \anynetxv{C} & + $g_{i+1} = g_i$   & 10 & $(16\c128)^4\c3\c6$        & \app$3.2\c10^{14}$\\
 \anynetxv{D} & + $w_{i+1} \ge w_i$ & 10 & $(16\c128)^4\c3\c6/(4!)$   & \app$1.3\c10^{13}$\\
 \anynetxv{E} & + $d_{i+1} \ge d_i$ & 10 & $(16\c128)^4\c3\c6/(4!)^2$ & \app$5.5\c10^{11}$\\
 \regnet      & quantized linear    & 6  & \app$64^4\c6\c3$           & \app$3.0\c10^{8}$\\
\end{tabular}}\vspace{1mm}
\caption{\textbf{Design space summary.} See text for details.}
\label{tab:sizes}
\end{table}

Table~\ref{tab:sizes} shows a summary of the design space sizes (for \regnet we estimate the size by quantizing its continuous parameters). In designing \regnetx, we reduced the dimension of the original \anynetx design space from 16 to 6 dimensions, and the size nearly 10 orders of magnitude. We note, however, that \regnet still contains a good diversity of models that can be tuned for a variety of settings.

\subsection{Design Space Generalization}\label{sec:dds:generalize}

\begin{figure}[t]\centering
\includegraphics[width=\linewidth]{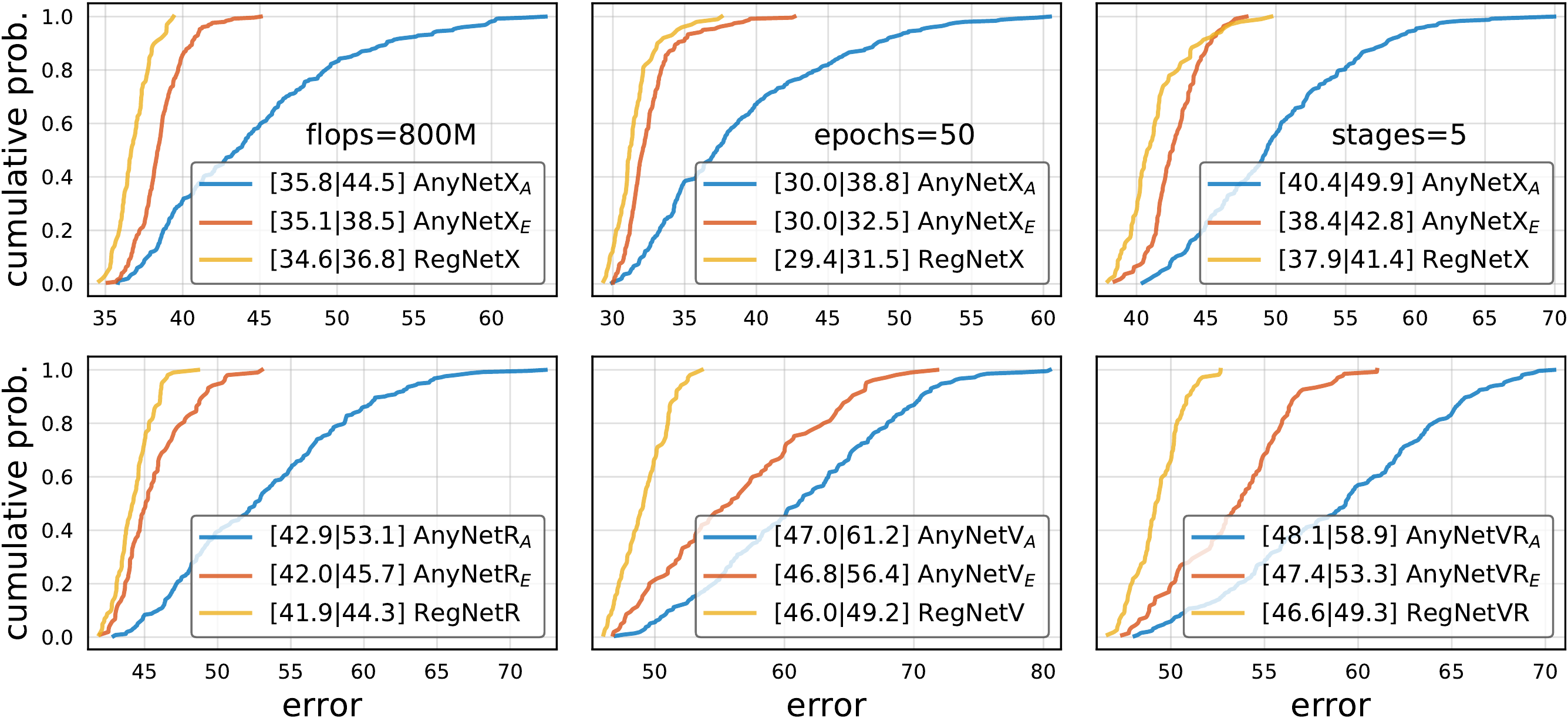}
\caption{\textbf{\regnetx generalization}. We compare \regnetx to \anynetx at higher flops (top-left), higher epochs (top-middle), with 5-stage networks (top-right), and with various block types (bottom). In all cases the ordering of the design spaces is consistent and we see no signs of design space overfitting.}
\label{fig:regnetx:generalize}\vspace{-3mm}
\end{figure}

We designed the \regnet design space in a low-compute, low-epoch training regime with only a single block type. However, our goal is not to design a design space for a single setting, but rather to discover general principles of network design that can \emph{generalize} to new settings.

In Figure~\ref{fig:regnetx:generalize}, we compare the \regnetx design space to \anynetxv{A} and \anynetxv{E} at higher flops, higher epochs, with 5-stage networks, and with various block types (described in the appendix). In all cases the ordering of the design spaces is consistent, with \regnetx $>$ \anynetxv{E} $>$ \anynetxv{A}. In other words, we see no signs of overfitting. These results are promising because they show \regnet can generalize to new settings. The 5-stage results show the regular structure of \regnet can generalize to more stages, where \anynetxv{A} has even more degrees of freedom.

\section{Analyzing the \dsnamebold{RegNetX} Design Space}\label{sec:regnet}

We next further analyze the \regnetx design space and revisit common deep network design choices. Our analysis yields surprising insights that don't match popular practice, which allows us to achieve good results with simple models.

As the \regnetx design space has a high concentration of good models, for the following results we switch to sampling fewer models (100) but training them for longer (25 epochs) with a learning rate of 0.1 (see appendix). We do so to observe more fine-grained trends in network behavior.

\begin{figure}[t]\centering
\includegraphics[width=\linewidth]{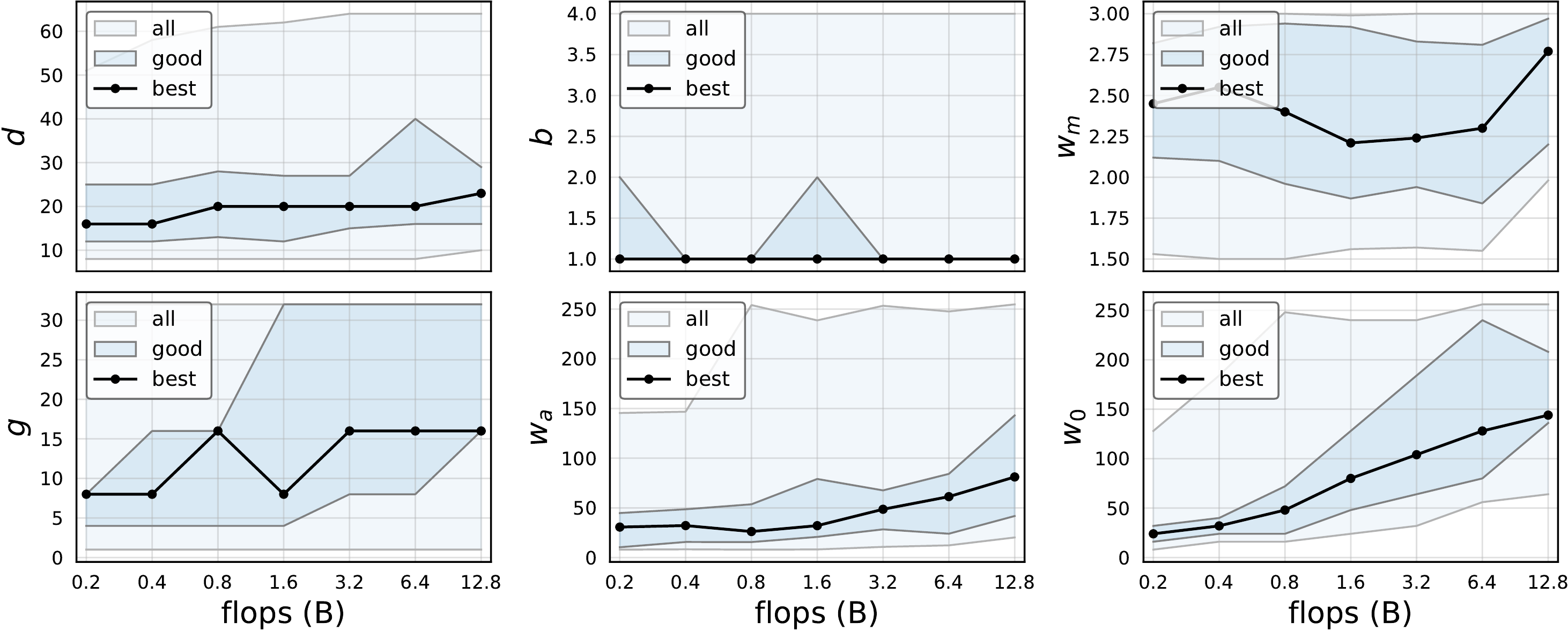}
\caption{\textbf{\regnetx parameter trends}. For each parameter and each flop regime we apply an empirical bootstrap to obtain the range that contains best models with 95\% confidence (shown with blue shading) and the likely best model (black line), see also Figure~\ref{fig:anynetx:a}. We observe that for best models the depths $d$ are remarkably stable across flops regimes, and $b=1$ and $w_m\approx2.5$ are best. Block and groups widths ($w_a$, $w_0$, $g$) tend to increase with flops.}
\label{fig:regnetx:trends}
\end{figure}

\begin{figure}[t]\centering
\begin{minipage}[t]{0.32\linewidth}\resizebox{\columnwidth}{!}{
\tablestyle{2pt}{1.3}
\begin{tabular}[b]{@{}l|ccc@{}}
  & flops & params & acts. \\\shline
 1$\x$1 conv       & $w^2r^2$    & $w^2$    & $wr^2$\\
 3$\x$3 conv       & $3^2w^2r^2$ & $3^2w^2$ & $wr^2$\\
 3$\x$3 gr conv    & $3^2wgr^2$  & $3^2wg$  & $wr^2$\\
 3$\x$3 dw conv    & $3^2wr^2$   & $3^2w$   & $wr^2$\\
 \multicolumn{3}{c}{~}\\ \multicolumn{3}{c}{~}\\
\end{tabular}}\end{minipage}\hspace{.5mm}
\includegraphics[width=.66\linewidth]{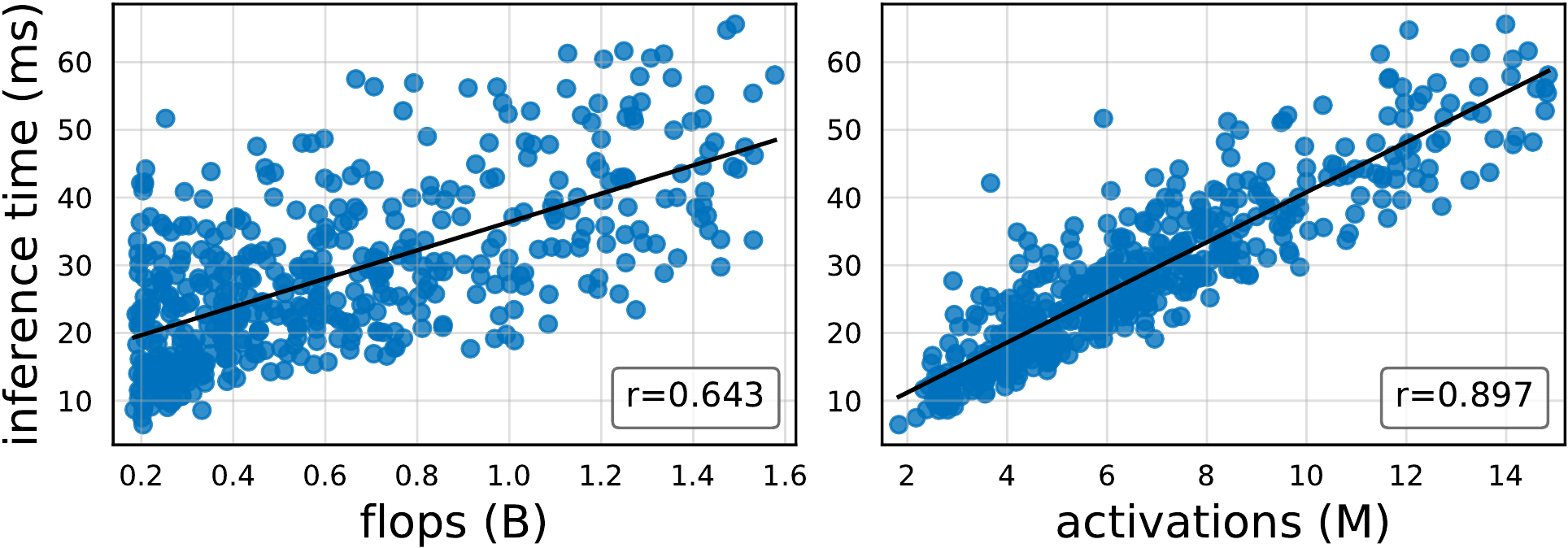}\\
\includegraphics[width=1.0\linewidth]{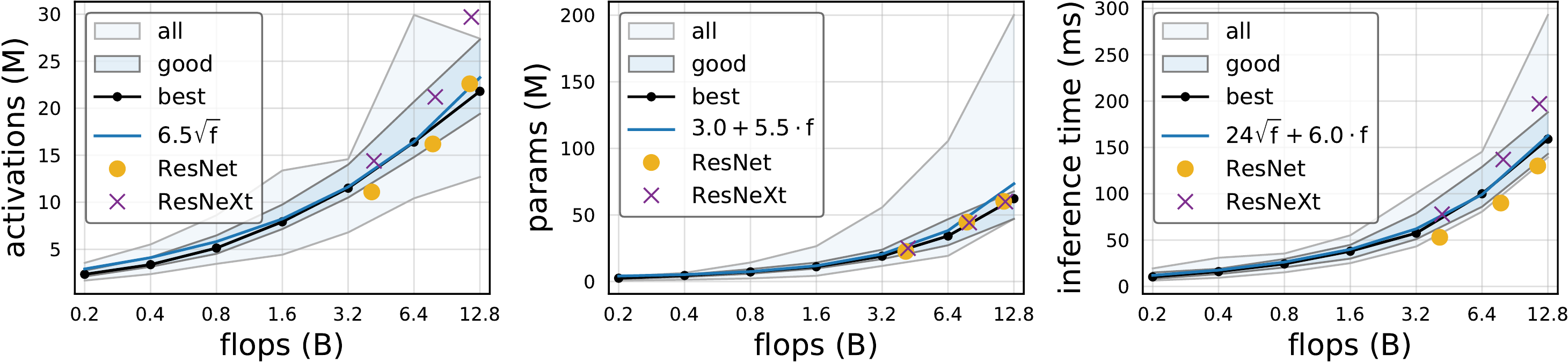}
\caption{\textbf{Complexity metrics}. \textit{Top:} Activations can have a stronger correlation to runtime on hardware accelerators than flops (we measure inference time for 64 images on an NVIDIA V100 GPU). \textit{Bottom:} Trend analysis of complexity \vs flops and best fit curves (shown in blue) of the trends for best models (black curves).}
\label{fig:regnetx:complexity}\vspace{-3mm}
\end{figure}

\paragraph{\dsnamebold{RegNet} trends.} We show trends in the \regnetx parameters across flop regimes in Figure~\ref{fig:regnetx:trends}. Remarkably, the \emph{depth of best models is stable across regimes} (top-left), with an optimal depth of \app20 blocks (60 layers). This is in contrast to the common practice of using deeper models for higher flop regimes. We also observe that \emph{the best models use a bottleneck ratio $b$ of 1.0} (top-middle), which effectively removes the bottleneck (commonly used in practice). Next, we observe that the width multiplier $w_m$ of good models is $\app2.5$ (top-right), similar but not identical to the popular recipe of doubling widths across stages. The remaining parameters ($g$, $w_a$, $w_0$) increase with complexity (bottom).

\paragraph{Complexity analysis.} In addition to flops and parameters, we analyze network \emph{activations}, which we define as the size of the output tensors of all conv layers (we list complexity measures of common conv operators in Figure~\ref{fig:regnetx:complexity}, top-left). While not a common measure of network complexity, activations can heavily affect runtime on memory-bound hardware accelerators (\eg, GPUs, TPUs), for example, see Figure~\ref{fig:regnetx:complexity} (top). In Figure~\ref{fig:regnetx:complexity} (bottom), we observe that for the best models in the population, activations increase with the square-root of flops, parameters increase linearly, and runtime is best modeled using both a linear and a square-root term due to its dependence on both flops and activations.

\begin{figure}[t]\centering
\includegraphics[height=19.5mm]{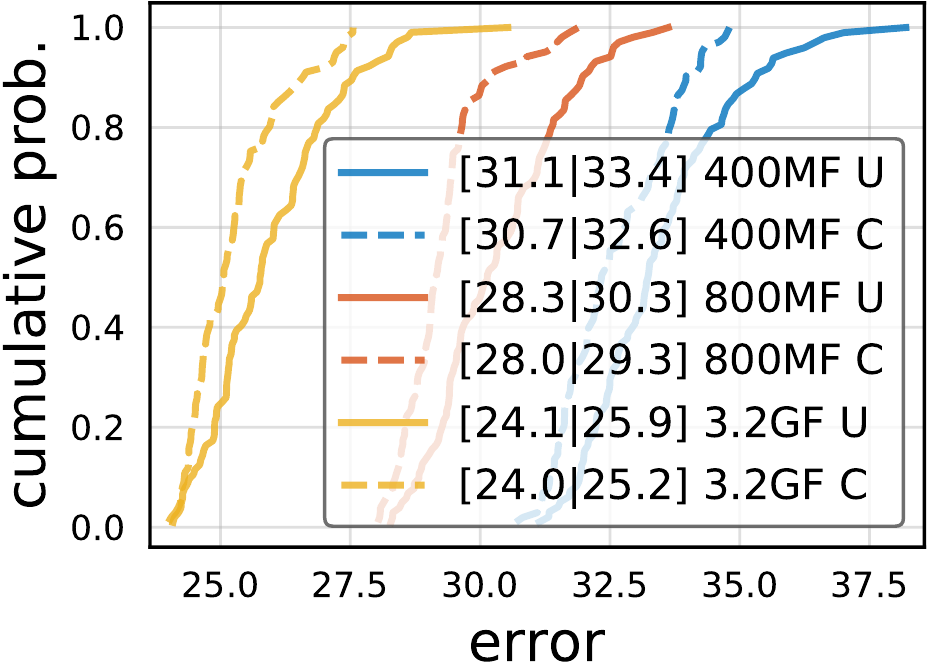}
\includegraphics[height=19.5mm]{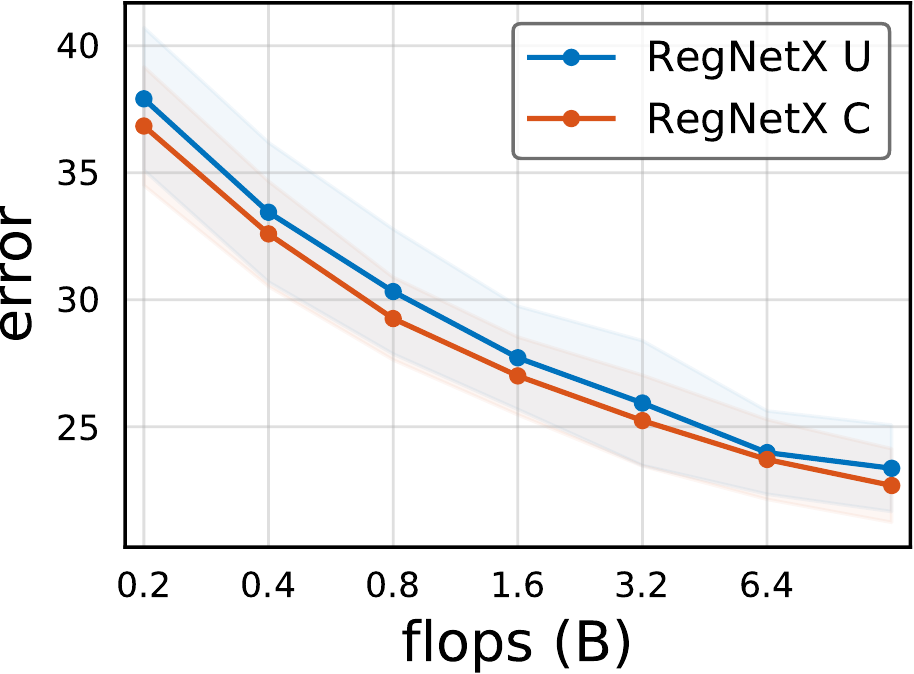}
\includegraphics[height=19.5mm]{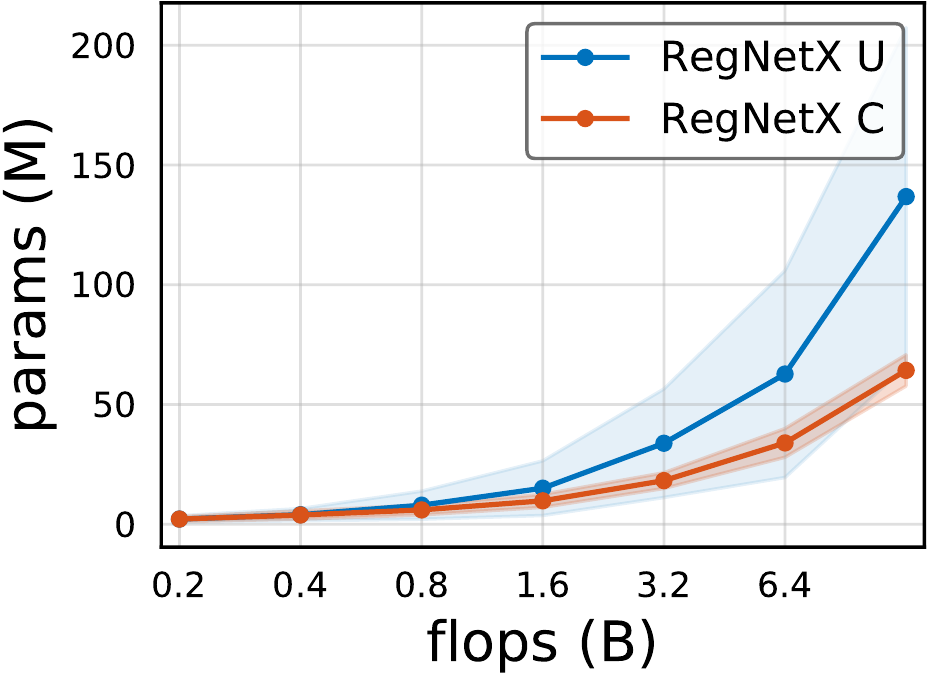}
\caption{We \textbf{refine \regnetx} using various constraints (see text). The constrained variant (C) is best across all flop regimes while being more efficient in terms of parameters and activations.}
\label{fig:regnetx:constrained}\vspace{-2mm}
\end{figure}

\paragraph{\regnetx constrained.} Using these findings, we refine the \regnetx design space. First, based on Figure~\ref{fig:regnetx:trends} (top), we set $b=1$, $d \le 40$, and $w_m \ge 2$. Second, we limit parameters and activations, following Figure~\ref{fig:regnetx:complexity} (bottom). This yields fast, low-parameter, low-memory models without affecting accuracy. In Figure~\ref{fig:regnetx:constrained}, we test \regnetx with theses constraints and observe that the constrained version is superior across all flop regimes. We use this version in \S\ref{sec:soa}, and further limit depth to $12 \le d \le 28$ (see also Appendix D).

\begin{figure}[t]\centering
\includegraphics[height=21mm]{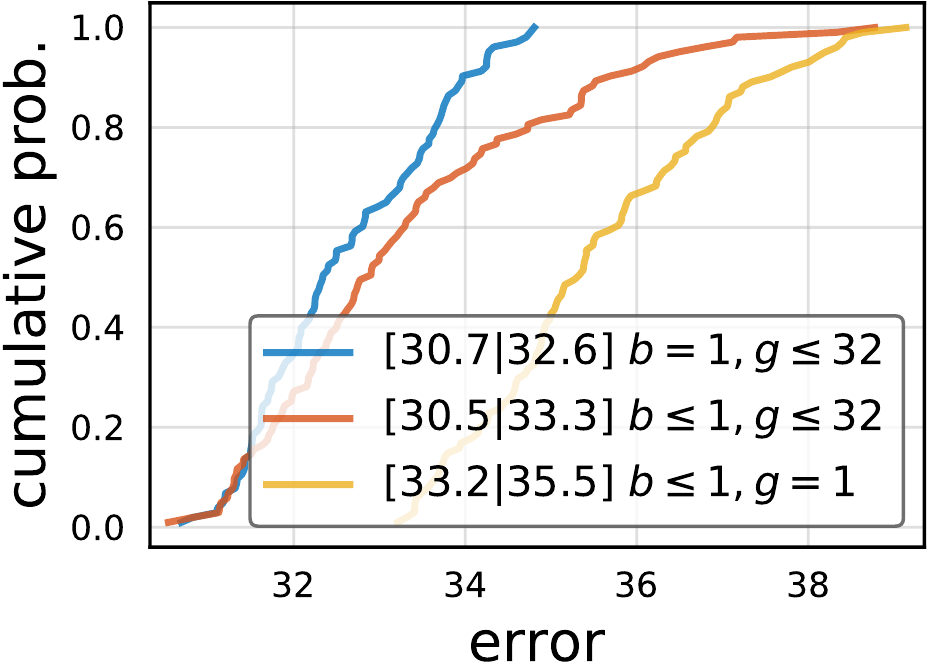}\hspace{.5mm}
\includegraphicstriml{40}{height=21mm}{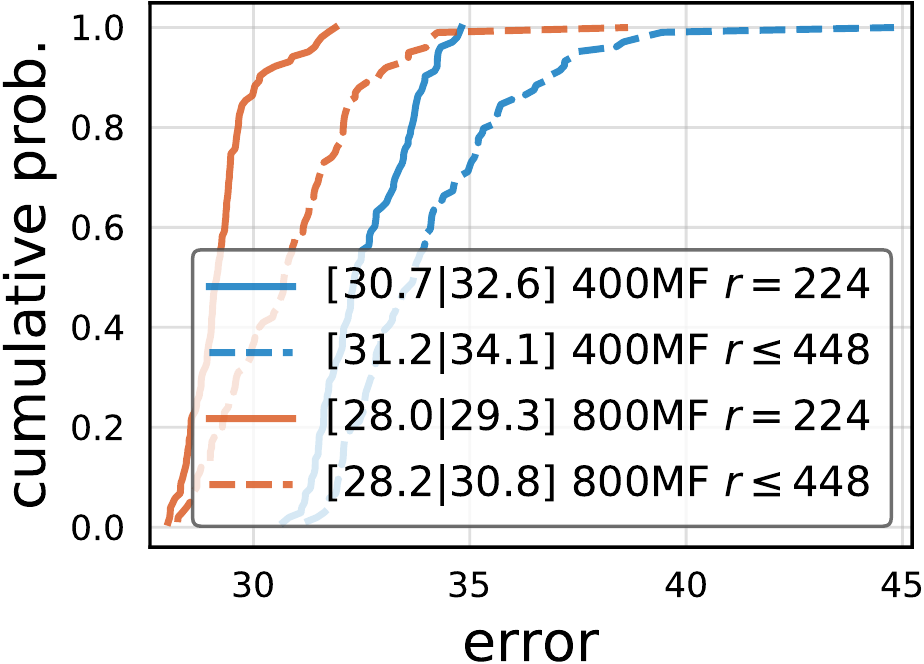}
\includegraphicstriml{40}{height=21mm}{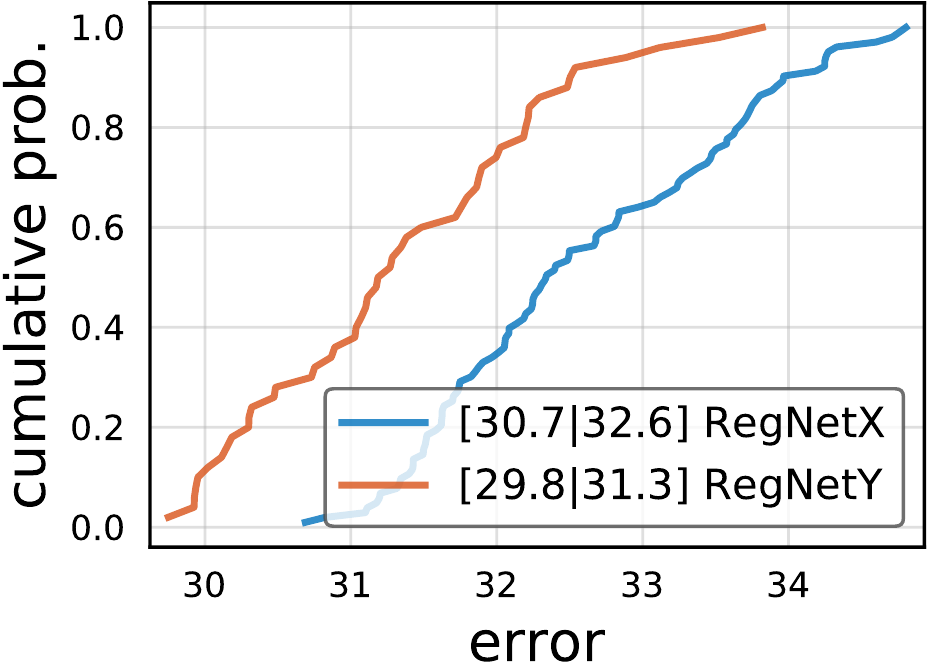}\\[.5mm]
\caption{We evaluate \regnetx with \textbf{alternate design choices}. \textit{Left:} Inverted bottleneck ($\frac{1}{8} \le b \le 1$) degrades results and depthwise conv ($g=1$) is even worse. \textit{Middle:} Varying resolution $r$ harms results. \textit{Right:} \regnety (\dsname{Y=X+SE}) improves the EDF.}
\label{fig:regnetx:alternatives}\vspace{-5mm}
\end{figure}

\paragraph{Alternate design choices.} Modern mobile networks often employ the inverted bottleneck ($b<1$) proposed in~\cite{Sandler2018} along with depthwise conv~\cite{Chollet2017} ($g=1$). In Figure~\ref{fig:regnetx:alternatives} (left), we observe that the \emph{inverted bottleneck degrades the EDF slightly and depthwise conv performs even worse} relative to $b=1$ and $g\ge1$ (see appendix for further analysis). Next, motivated by \cite{Tan2019} who found that scaling the input image resolution can be helpful, we test varying resolution in Figure~\ref{fig:regnetx:alternatives} (middle). Contrary to \cite{Tan2019}, we find that for \regnetx a fixed resolution of $224\x224$ is best, even at higher flops.

\paragraph{SE.} Finally, we evaluate \regnetx with the popular Squeeze-and-Excitation (SE) op~\cite{Hu2018} (we abbreviate \dsname{X+SE} as \dsname{Y} and refer to the resulting design space as \regnety). In Figure~\ref{fig:regnetx:alternatives} (right), we see that \regnety yields good gains.

\section{Comparison to Existing Networks}\label{sec:soa}

We now compare top models from the \regnetx and \regnety design spaces at various complexities to the state-of-the-art on ImageNet~\cite{Deng2009}. We denote individual models using small caps, \eg \model{RegNetX}. We also suffix the models with the flop regime, \eg 400MF. For each flop regime, we pick the best model from 25 random settings of the \regnet parameters ($d$, $g$, $w_m$, $w_a$, $w_0$), and re-train the top model 5 times at 100 epochs to obtain robust error estimates.

\begin{figure}[t]\centering
\includegraphics[width=\linewidth]{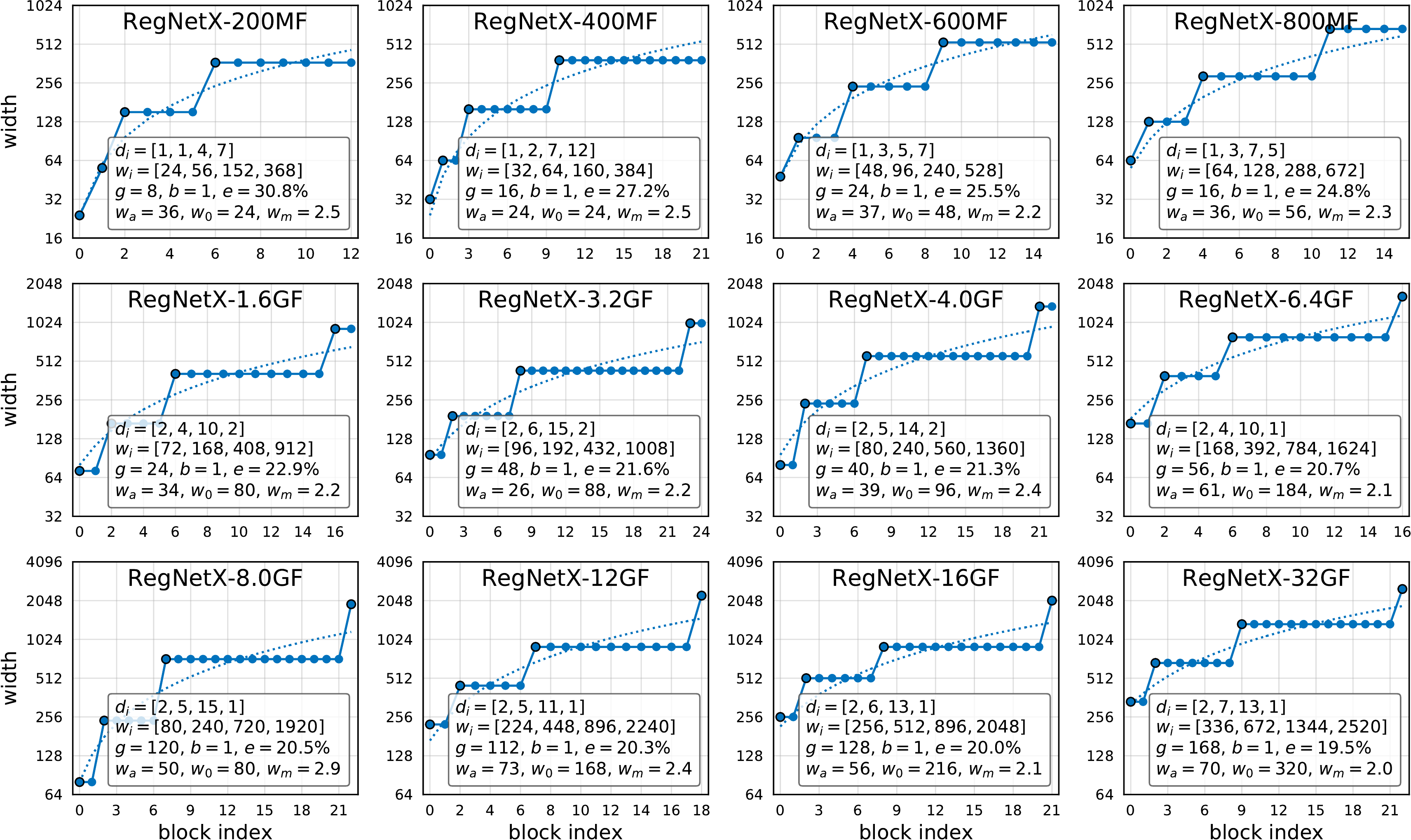}\vspace{2mm}
\resizebox{\columnwidth}{!}{\tablestyle{8pt}{1.05}
\begin{tabular}{@{}m|ccc|ccc|c@{}}
  & flops & params & acts & batch & infer & train  & error\\
  & (B)   & (M)    & (M)  & size  & (ms)  & (hr)   & (top-1)\\\shline
 RegNetX-200MF  &  0.2 &   2.7 &  2.2 & 1024 &  10 &   2.8 & 31.1\mypm{0.09} \\
 RegNetX-400MF  &  0.4 &   5.2 &  3.1 & 1024 &  15 &   3.9 & 27.3\mypm{0.15} \\
 RegNetX-600MF  &  0.6 &   6.2 &  4.0 & 1024 &  17 &   4.4 & 25.9\mypm{0.03} \\
 RegNetX-800MF  &  0.8 &   7.3 &  5.1 & 1024 &  21 &   5.7 & 24.8\mypm{0.09} \\
 RegNetX-1.6GF  &  1.6 &   9.2 &  7.9 & 1024 &  33 &   8.7 & 23.0\mypm{0.13} \\
 RegNetX-3.2GF  &  3.2 &  15.3 & 11.4 &  512 &  57 &  14.3 & 21.7\mypm{0.08} \\
 RegNetX-4.0GF  &  4.0 &  22.1 & 12.2 &  512 &  69 &  17.1 & 21.4\mypm{0.19} \\
 RegNetX-6.4GF  &  6.5 &  26.2 & 16.4 &  512 &  92 &  23.5 & 20.8\mypm{0.07} \\
 RegNetX-8.0GF  &  8.0 &  39.6 & 14.1 &  512 &  94 &  22.6 & 20.7\mypm{0.07} \\
 RegNetX-12GF   & 12.1 &  46.1 & 21.4 &  512 & 137 &  32.9 & 20.3\mypm{0.04} \\
 RegNetX-16GF   & 15.9 &  54.3 & 25.5 &  512 & 168 &  39.7 & 20.0\mypm{0.11} \\
 RegNetX-32GF   & 31.7 & 107.8 & 36.3 &  256 & 318 &  76.9 & 19.5\mypm{0.12} \\
\end{tabular}}\vspace{1mm}
\caption{\textbf{Top \model{RegNetX} models}. We measure inference time for 64 images on an NVIDIA V100 GPU; train time is for 100 epochs on 8 GPUs with the batch size listed. Network diagram legends contain all information required to implement the models.}
\label{fig:models:regnetx}\vspace{-4mm}
\end{figure}

\begin{figure}[t]\centering
\includegraphics[width=\linewidth]{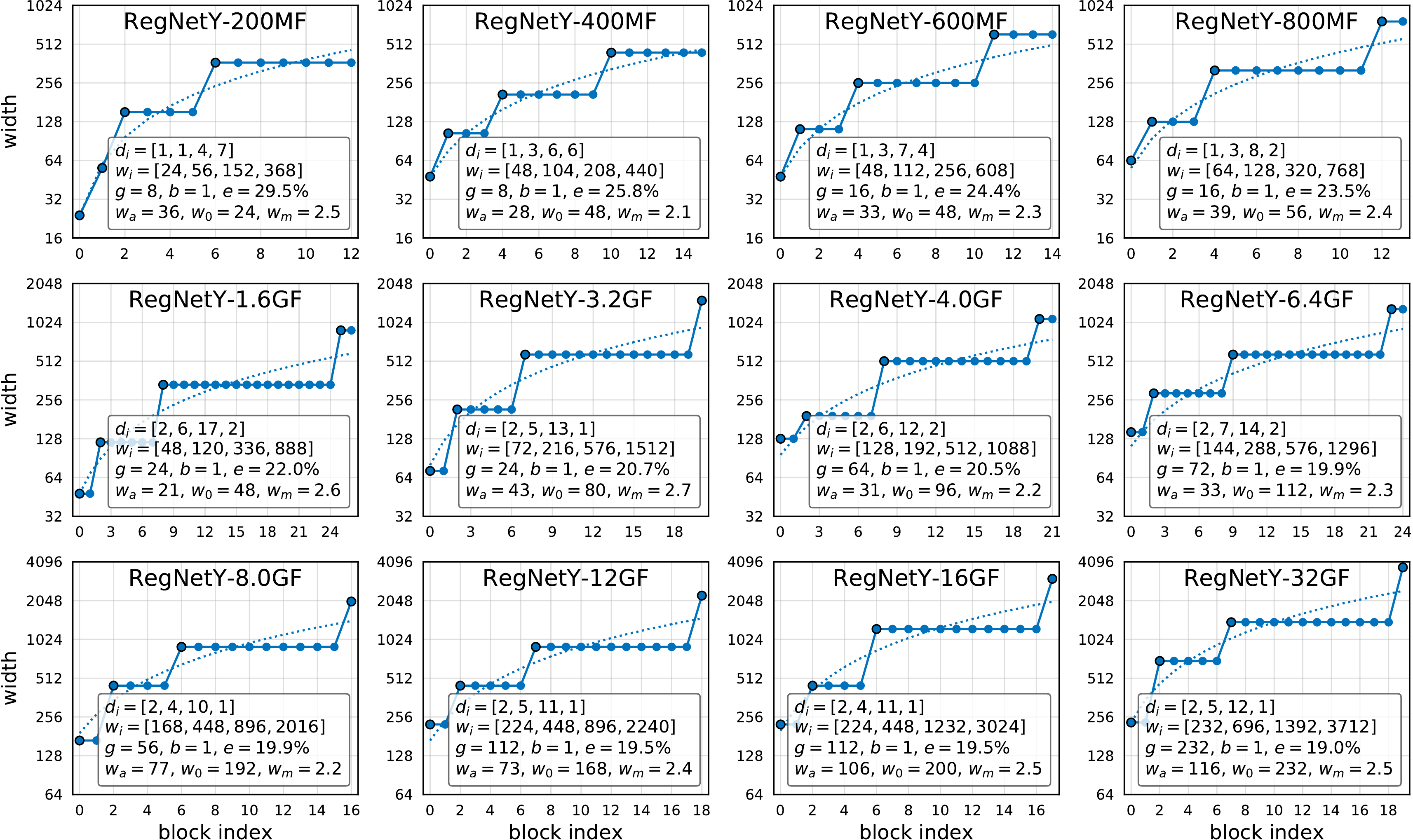}\vspace{2mm}
\resizebox{\columnwidth}{!}{\tablestyle{8pt}{1.05}
\begin{tabular}{@{}m|ccc|ccc|c@{}}
  & flops & params & acts & batch & infer & train  & error\\
  & (B)   & (M)    & (M)  & size  & (ms)  & (hr)   & (top-1)\\\shline
 RegNetY-200MF  &  0.2 &   3.2 &  2.2 & 1024 &  11 &   3.1 & 29.6\mypm{0.11} \\
 RegNetY-400MF  &  0.4 &   4.3 &  3.9 & 1024 &  19 &   5.1 & 25.9\mypm{0.16} \\
 RegNetY-600MF  &  0.6 &   6.1 &  4.3 & 1024 &  19 &   5.2 & 24.5\mypm{0.07} \\
 RegNetY-800MF  &  0.8 &   6.3 &  5.2 & 1024 &  22 &   6.0 & 23.7\mypm{0.03} \\
 RegNetY-1.6GF  &  1.6 &  11.2 &  8.0 & 1024 &  39 &  10.1 & 22.0\mypm{0.08} \\
 RegNetY-3.2GF  &  3.2 &  19.4 & 11.3 &  512 &  67 &  16.5 & 21.0\mypm{0.05} \\
 RegNetY-4.0GF  &  4.0 &  20.6 & 12.3 &  512 &  68 &  16.8 & 20.6\mypm{0.08} \\
 RegNetY-6.4GF  &  6.4 &  30.6 & 16.4 &  512 & 104 &  26.1 & 20.1\mypm{0.04} \\
 RegNetY-8.0GF  &  8.0 &  39.2 & 18.0 &  512 & 113 &  28.1 & 20.1\mypm{0.09} \\
 RegNetY-12GF   & 12.1 &  51.8 & 21.4 &  512 & 150 &  36.0 & 19.7\mypm{0.06} \\
 RegNetY-16GF   & 15.9 &  83.6 & 23.0 &  512 & 189 &  45.6 & 19.6\mypm{0.16} \\
 RegNetY-32GF   & 32.3 & 145.0 & 30.3 &  256 & 319 &  76.0 & 19.0\mypm{0.12} \\
\end{tabular}}\vspace{1mm}
\caption{\textbf{Top \model{RegNetY} models} (\dsname{Y=X+SE}). The benchmarking setup and the figure format is the same as in Figure~\ref{fig:models:regnetx}.}
\label{fig:models:regnety}\vspace{-4mm}
\end{figure}

Resulting top \model{RegNetX} and \model{RegNetY} models for each flop regime are shown in Figures~\ref{fig:models:regnetx} and \ref{fig:models:regnety}, respectively. In addition to the simple linear structure and the trends we analyzed in \S\ref{sec:regnet}, we observe an interesting pattern. Namely, the higher flop models have a large number of blocks in the third stage and a small number of blocks in the last stage. This is similar to the design of standard \model{ResNet} models. Moreover, we observe that the group width $g$ increases with complexity, but depth $d$ saturates for large models.

Our goal is to perform fair comparisons and provide simple and easy-to-reproduce baselines. We note that along with better architectures, much of the recently reported gains in network performance are based on enhancements to the \emph{training setup and regularization scheme} (see Table~\ref{tab:optimization:en:b0}). As our focus is on evaluating \emph{network architectures}, we perform carefully controlled experiments under the same training setup. In particular, to provide fair comparisons to classic work, \emph{we do not use any training-time enhancements}.

\subsection{State-of-the-Art Comparison: Mobile Regime}\label{sec:soa:mobile}

Much of the recent work on network design has focused on the mobile regime (\app600MF). In Table~\ref{tab:comparison:mobile}, we compare \model{RegNet} models at 600MF to existing mobile networks. We observe that \model{RegNets} are surprisingly effective in this regime considering the substantial body of work on finding better mobile networks via both manual design~\cite{Howard2017, Sandler2018, Ma2018} and NAS ~\cite{Zoph2018, Real2018, Liu2018, Liu2019}.

We emphasize that \model{RegNet} models use our basic 100 epoch schedule with no regularization except weight decay, while most mobile networks use longer schedules with various enhancements, such as deep supervision~\cite{Lee2015}, Cutout~\cite{Devries2017}, DropPath~\cite{Larsson2017}, AutoAugment~\cite{Cubuk2018}, and so on. As such, we hope our strong results obtained with a short training schedule without enhancements can serve as a simple baseline for future work.

\begin{table}[h]\centering\vspace{-1mm}
\resizebox{.9\columnwidth}{!}{\tablestyle{10pt}{1.05}
\begin{tabular}{@{}m|cc|l@{}}
 & flops (B) & params (M) & top-1 error\\\shline
 MobileNet~\cite{Howard2017}     & 0.57  & 4.2  & 29.4 \\
 MobileNet-V2~\cite{Sandler2018} & 0.59  & 6.9  & 25.3 \\
 ShuffleNet~\cite{Zhang2018}     & 0.52  &  -   & 26.3 \\
 ShuffleNet-V2~\cite{Ma2018}     & 0.59  &  -   & 25.1 \\
 NASNet-A~\cite{Zoph2018}        & 0.56  & 5.3  & 26.0 \\
 AmoebaNet-C~\cite{Real2018}     & 0.57  & 6.4  & 24.3 \\
 PNASNet-5~\cite{Liu2018}        & 0.59  & 5.1  & 25.8 \\
 DARTS~\cite{Liu2019}            & 0.57  & 4.7  & 26.7 \\\hline
 RegNetX-600MF                   & 0.60  & 6.2  & 25.9\mypm{0.03} \\
 RegNetY-600MF                   & 0.60  & 6.1  & 24.5\mypm{0.07} \\
\end{tabular}}\vspace{1.5mm}
\caption{\textbf{Mobile regime.} We compare existing models using \emph{originally} reported errors to \regnet models trained in a \emph{basic} setup. Our \emph{simple} \regnet models achieve surprisingly good results given the effort focused on this regime in the past few years.}
\label{tab:comparison:mobile}\vspace{-2mm}
\end{table}

\begin{figure}[t]\centering
\includegraphics[width=1.0\linewidth]{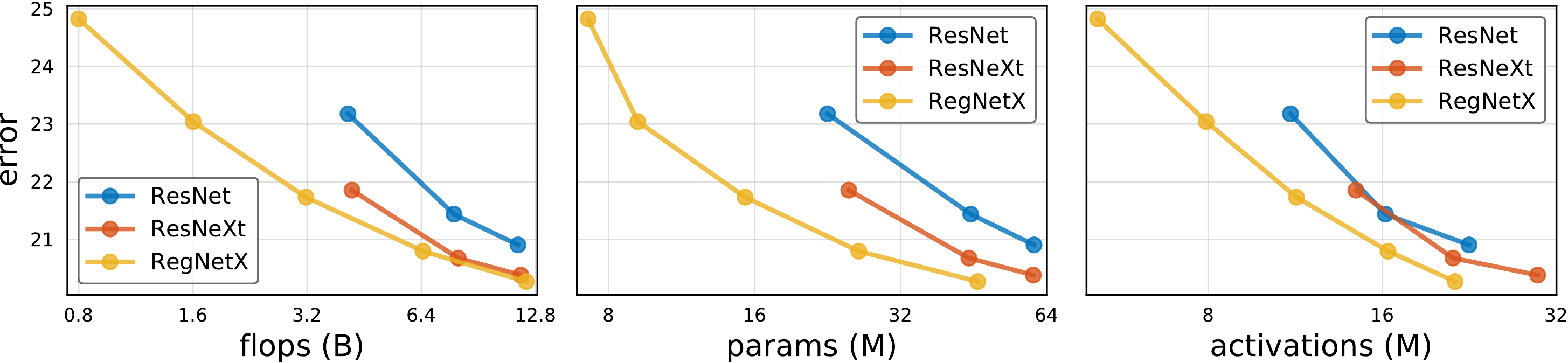}
\caption{\textbf{ResNe(X)t comparisons.} \model{RegNetX} models versus \model{ResNe(X)t-(50,101,152)} under various complexity metrics. As all models use the identical components and training settings, all observed gains are from the design of the \regnetx design space.}
\label{fig:comparison:resnet}\vspace{-0mm}
\end{figure}

\begin{table}[t]\centering
\resizebox{\columnwidth}{!}{\tablestyle{10pt}{1.05}
\begin{tabular}{@{}m|ccc|cc|l@{}}
  & flops & params & acts & infer & train  & top-1 error\\
  & (B)   & (M)    & (M)  & (ms)  & (hr)   & ours\mypm{std}~~~\origerr{orig}\\\shline
 ResNet-50      & \g  4.1 & \g 22.6 & 11.1 &  53 & 12.2 &     {23.2}\mypm{0.09} \origerr{23.9} \\
 RegNetX-3.2GF  & \g  3.2 & \g 15.3 & 11.4 &  57 & 14.3 & {\bf 21.7}\mypm{0.08} \\\hline
 ResNeXt-50     & \g  4.2 & \g 25.0 & 14.4 &  78 & 18.0 &     {21.9}\mypm{0.10} \origerr{22.2} \\
 ResNet-101     & \g  7.8 & \g 44.6 & 16.2 &  90 & 20.4 &     {21.4}\mypm{0.11} \origerr{22.0} \\
 RegNetX-6.4GF  & \g  6.5 & \g 26.2 & 16.4 &  92 & 23.5 & {\bf 20.8}\mypm{0.07} \\\hline
 ResNeXt-101    & \g  8.0 & \g 44.2 & 21.2 & 137 & 31.8 &     {20.7}\mypm{0.08} \origerr{21.2}\\
 ResNet-152     & \g 11.5 & \g 60.2 & 22.6 & 130 & 29.2 &     {20.9}\mypm{0.12} \origerr{21.6}\\
 RegNetX-12GF   & \g 12.1 & \g 46.1 & 21.4 & 137 & 32.9 & {\bf 20.3}\mypm{0.04} \\
 \multicolumn{7}{c}{(a) Comparisons grouped by \textbf{activations}.} \vspace{2mm}\\\shline
 ResNet-50      &  4.1 & 22.6 & \g 11.1 &  53 & 12.2 &     {23.2}\mypm{0.09} \origerr{23.9} \\
 ResNeXt-50     &  4.2 & 25.0 & \g 14.4 &  78 & 18.0 &     {21.9}\mypm{0.10} \origerr{22.2} \\
 RegNetX-4.0GF  &  4.0 & 22.1 & \g 12.2 &  69 & 17.1 & {\bf 21.4}\mypm{0.19} \\\hline
 ResNet-101     &  7.8 & 44.6 & \g 16.2 &  90 & 20.4 &     {21.4}\mypm{0.11} \origerr{22.0} \\
 ResNeXt-101    &  8.0 & 44.2 & \g 21.2 & 137 & 31.8 & {\bf 20.7}\mypm{0.08} \origerr{21.2} \\
 RegNetX-8.0GF  &  8.0 & 39.6 & \g 14.1 &  94 & 22.6 & {\bf 20.7}\mypm{0.07} \\\hline
 ResNet-152     & 11.5 & 60.2 & \g 22.6 & 130 & 29.2 &     {20.9}\mypm{0.12} \origerr{21.6} \\
 ResNeXt-152    & 11.7 & 60.0 & \g 29.7 & 197 & 45.7 &     {20.4}\mypm{0.06} \origerr{21.1} \\
 RegNetX-12GF   & 12.1 & 46.1 & \g 21.4 & 137 & 32.9 & {\bf 20.3}\mypm{0.04} \\
 \multicolumn{7}{c}{(b) Comparisons grouped by \textbf{flops}.} \\
\end{tabular}}\vspace{1.5mm}
\caption{\textbf{\model{ResNe(X)t} comparisons.} (a) Grouped by activations, \model{RegNetX} show considerable gains (note that for each group GPU inference and training times are similar). (b) \model{RegNetX} models outperform \model{ResNe(X)t} models under fixed flops as well.}
\label{tab:comparison:resnet}\vspace{-3mm}
\end{table}

\subsection{Standard Baselines Comparison: ResNe(X)t}\label{sec:soa:resnext}

Next, we compare \model{RegNetX} to standard \model{ResNet}~\cite{He2016} and \model{ResNeXt}~\cite{Xie2017} models. All of the models in this experiment come from the exact same design space, the former being manually designed, the latter being obtained through design space design.  For fair comparisons, we compare \model{RegNet} and \model{ResNe(X)t} models under the same training setup (our standard \model{RegNet} training setup). We note that this results in improved \model{ResNe(X)t} baselines and highlights the importance of carefully controlling the training setup.

Comparisons are shown in Figure~\ref{fig:comparison:resnet} and Table~\ref{tab:comparison:resnet}. Overall, we see that \model{RegNetX} models, by optimizing the network structure alone, provide considerable improvements under all complexity metrics. We emphasize that good \model{RegNet} models are available across a wide range of compute regimes, including in low-compute regimes where good \model{ResNe(X)t} models are not available. 

Table~\ref{tab:comparison:resnet}a shows comparisons grouped by activations (which can strongly influence runtime on accelerators such as GPUs). This setting is of particular interest to the research community where model training time is a bottleneck and will likely have more real-world use cases in the future, especially as accelerators gain more use at inference time (\eg, in self-driving cars). \model{RegNetX} models are quite effective given a fixed inference or training time budget.

\begin{figure}[t]\centering
\includegraphics[width=1.0\linewidth]{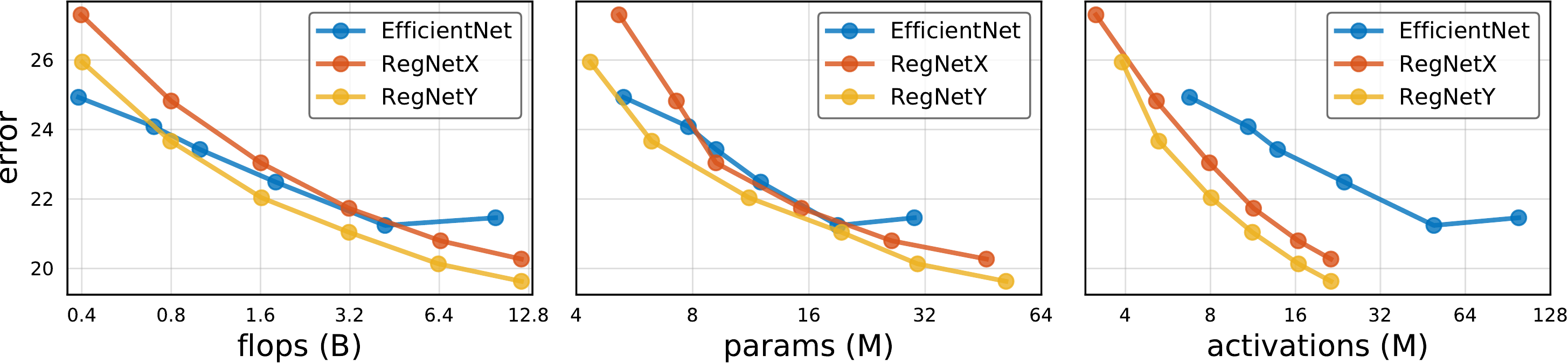}
\caption{\textbf{\model{EfficientNet} comparisons.} \model{RegNet}s outperform the state of the art, especially when considering activations.}
\label{fig:comparison:en}\vspace{-2mm}
\end{figure}

\begin{table}[t]\centering
\resizebox{\columnwidth}{!}{\tablestyle{6pt}{1.05}
\begin{tabular}{@{}m|ccc|ccc|l@{}}
  & flops & params & acts & batch & infer & train  & top-1 error\\
  & (B)   & (M)    & (M)  & size  & (ms)  & (hr)   & ours\mypm{std}~~~\origerr{orig}\\\shline
 EfficientNet-B0 & 0.4  &  5.3 &      6.7 &  256 &  34 &  11.7 & {\bf 24.9}\mypm{0.03} \origerr{23.7} \\
 RegNetY-400MF   & 0.4  &  4.3 & \bf  3.9 & 1024 &  19 &   5.1 & 25.9\mypm{0.16} \\\hline
 EfficientNet-B1 & 0.7  &  7.8 &     10.9 &  256 &  52 &  15.6 & {\bf 24.1}\mypm{0.16} \origerr{21.2} \\
 RegNetY-600MF   & 0.6  &  6.1 & \bf  4.3 & 1024 &  19 &   5.2 &     {24.5}\mypm{0.07} \\\hline
 EfficientNet-B2 & 1.0  &  9.2 &     13.8 &  256 &  68 &  18.4 & {\bf 23.4}\mypm{0.06} \origerr{20.2} \\
 RegNetY-800MF   & 0.8  &  6.3 & \bf  5.2 & 1024 &  22 &   6.0 &     {23.7}\mypm{0.03} \\\hline
 EfficientNet-B3 & 1.8  & 12.0 &     23.8 &  256 & 114 &  32.1 &     {22.5}\mypm{0.05} \origerr{18.9} \\
 RegNetY-1.6GF   & 1.6  & 11.2 & \bf  8.0 & 1024 &  39 &  10.1 & {\bf 22.0}\mypm{0.08} \\\hline
 EfficientNet-B4 & 4.2  & 19.0 &     48.5 &  128 & 240 &  65.1 &     {21.2}\mypm{0.06} \origerr{17.4} \\
 RegNetY-4.0GF   & 4.0  & 20.6 & \bf 12.3 &  512 &  68 &  16.8 & {\bf 20.6}\mypm{0.08} \\\hline
 EfficientNet-B5 & 9.9  & 30.0 &     98.9 &   64 & 504 & 135.1 &     {21.5}\mypm{0.11} \origerr{16.7} \\
 RegNetY-8.0GF   & 8.0  & 39.2 & \bf 18.0 &  512 & 113 &  28.1 & {\bf 20.1}\mypm{0.09} \\
\end{tabular}}\vspace{1.5mm}
\caption{\textbf{\model{EfficientNet} comparisons} using our standard training schedule. Under \emph{comparable training settings}, \model{RegNetY} outperforms \model{EfficientNet} for most flop regimes. Moreover, \model{RegNet} models are considerably faster, \eg, \model{RegNetX-F8000} is about \emph{5$\x$ faster} than \model{EfficientNet-B5}. Note that originally reported errors for \model{EfficientNet} (shown grayed out), are much lower but use longer and enhanced training schedules, see Table~\ref{tab:optimization:en:b0}.}
\label{tab:comparison:en}\vspace{-4mm}
\end{table}

\subsection{State-of-the-Art Comparison: Full Regime}\label{sec:soa:EN}

We focus our comparison on \model{EfficientNet}~\cite{Tan2019}, which is representative of the state of the art and has reported impressive gains using a combination of NAS and an interesting model scaling rule across complexity regimes.

To enable direct comparisons, and to isolate gains due to improvements solely of the \emph{network architecture}, we opt to reproduce the exact \model{EfficientNet} models but using our standard training setup, with a 100 epoch schedule and no regularization except weight decay (effect of longer schedule and stronger regularization are shown in Table~\ref{tab:optimization:en:b0}). We optimize only $lr$ and $wd$, see Figure~\ref{fig:optimization:en} in appendix. This is the same setup as \model{RegNet} and enables fair comparisons.

Results are shown in Figure~\ref{fig:comparison:en} and Table~\ref{tab:comparison:en}. At low flops, \model{EfficientNet} outperforms the \model{RegNetY}. At intermediate flops, \model{RegNetY} outperforms \model{EfficientNet}, and at higher flops both \model{RegNetX} and \model{RegNetY} perform better.

We also observe that for \model{EfficientNet}, activations scale \emph{linearly} with flops (due to the scaling of both resolution and depth), compared to activations scaling with the \emph{square-root} of flops for \model{RegNet}s. This leads to slow GPU training and inference times for \model{EfficientNet}. \Eg, \model{RegNetX-8000} is \emph{5$\x$ faster} than \model{EfficientNet-B5}, while having lower error.

\section{Conclusion}

In this work, we present a new network design paradigm. Our results suggest that \emph{designing network design spaces} is a promising avenue for future research.

\section*{Appendix A: Test Set Evaluation}

\begin{table}[t]\centering
\resizebox{\columnwidth}{!}{\tablestyle{10pt}{1.05}
\begin{tabular}{@{}m|ccc|cc|l@{}}
  & flops & params & acts & infer & train  & error\\
  & (B)   & (M)    & (M)  & (ms)  & (hr)   & (top-1)\\\shline
 ResNet-50      & \g  4.1 & \g 22.6 & 11.1 &  53 & 12.2 &     {35.0}\mypm{0.20} \\
 RegNetX-3.2GF  & \g  3.2 & \g 15.3 & 11.4 &  57 & 14.3 & {\bf 33.6}\mypm{0.25} \\\hline
 ResNeXt-50     & \g  4.2 & \g 25.0 & 14.4 &  78 & 18.0 &     {33.5}\mypm{0.10} \\
 ResNet-101     & \g  7.8 & \g 44.6 & 16.2 &  90 & 20.4 &     {33.2}\mypm{0.24} \\
 RegNetX-6.4GF  & \g  6.5 & \g 26.2 & 16.4 &  92 & 23.5 & {\bf 32.6}\mypm{0.15} \\\hline
 ResNeXt-101    & \g  8.0 & \g 44.2 & 21.2 & 137 & 31.8 &     {32.1}\mypm{0.30} \\
 ResNet-152     & \g 11.5 & \g 60.2 & 22.6 & 130 & 29.2 &     {32.2}\mypm{0.22} \\
 RegNetX-12GF   & \g 12.1 & \g 46.1 & 21.4 & 137 & 32.9 & {\bf 32.0}\mypm{0.27} \\
 \multicolumn{7}{c}{(a) Comparisons grouped by \textbf{activations}.} \\[2mm]\shline
 ResNet-50      &  4.1 & 22.6 & \g 11.1 &  53 & 12.2 &     {35.0}\mypm{0.20} \\
 ResNeXt-50     &  4.2 & 25.0 & \g 14.4 &  78 & 18.0 &     {33.5}\mypm{0.10} \\
 RegNetX-4.0GF  &  4.0 & 22.1 & \g 12.2 &  69 & 17.1 & {\bf 33.2}\mypm{0.20} \\\hline
 ResNet-101     &  7.8 & 44.6 & \g 16.2 &  90 & 20.4 &     {33.2}\mypm{0.24} \\
 ResNeXt-101    &  8.0 & 44.2 & \g 21.2 & 137 & 31.8 & {\bf 32.1}\mypm{0.30} \\
 RegNetX-8.0GF  &  8.0 & 39.6 & \g 14.1 &  94 & 22.6 &     {32.5}\mypm{0.18} \\\hline
 ResNet-152     & 11.5 & 60.2 & \g 22.6 & 130 & 29.2 &     {32.2}\mypm{0.22} \\
 ResNeXt-152    & 11.7 & 60.0 & \g 29.7 & 197 & 45.7 & {\bf 31.5}\mypm{0.26} \\
 RegNetX-12GF   & 12.1 & 46.1 & \g 21.4 & 137 & 32.9 &     {32.0}\mypm{0.27} \\
 \multicolumn{7}{c}{(b) Comparisons grouped by \textbf{flops}.} \\
\end{tabular}}\vspace{1.5mm}
\caption{\textbf{\model{ResNe(X)t} comparisons on ImageNetV2.}}
\label{tab:comparison:resnet:v2}\vspace{-1mm}
\end{table}

\begin{table}[t]\centering
\resizebox{\columnwidth}{!}{\tablestyle{6pt}{1.05}
\begin{tabular}{@{}m|ccc|ccc|l@{}}
  & flops & params & acts & batch & infer & train  & error\\
  & (B)   & (M)    & (M)  & size  & (ms)  & (hr)   & (top-1)\\\shline
 EfficientNet-B0 & 0.4  &  5.3 &      6.7 &  256 &  34 &  11.7 & {\bf 37.1}\mypm{0.22} \\
 RegNetY-400MF   & 0.4  &  4.3 & \bf  3.9 & 1024 &  19 &   5.1 &     {38.3}\mypm{0.26} \\\hline
 EfficientNet-B1 & 0.7  &  7.8 &     10.9 &  256 &  52 &  15.6 & {\bf 36.4}\mypm{0.10} \\
 RegNetY-600MF   & 0.6  &  6.1 & \bf  4.3 & 1024 &  19 &   5.2 &     {36.9}\mypm{0.17} \\\hline
 EfficientNet-B2 & 1.0  &  9.2 &     13.8 &  256 &  68 &  18.4 & {\bf 35.3}\mypm{0.25} \\
 RegNetY-800MF   & 0.8  &  6.3 & \bf  5.2 & 1024 &  22 &   6.0 &     {35.7}\mypm{0.40} \\\hline
 EfficientNet-B3 & 1.8  & 12.0 &     23.8 &  256 & 114 &  32.1 &     {34.4}\mypm{0.27} \\
 RegNetY-1.6GF   & 1.6  & 11.2 & \bf  8.0 & 1024 &  39 &  10.1 & {\bf 33.9}\mypm{0.19} \\\hline
 EfficientNet-B4 & 4.2  & 19.0 &     48.5 &  128 & 240 &  65.1 &     {32.5}\mypm{0.23} \\
 RegNetY-4.0GF   & 4.0  & 20.6 & \bf 12.3 &  512 &  68 &  16.8 & {\bf 32.3}\mypm{0.28} \\\hline
 EfficientNet-B5 & 9.9  & 30.0 &     98.9 &   64 & 504 & 135.1 &     {31.5}\mypm{0.17} \\
 RegNetY-8.0GF   & 8.0  & 39.2 & \bf 18.0 &  512 & 113 &  28.1 & {\bf 31.3}\mypm{0.08} \\
\end{tabular}}\vspace{1.5mm}
\caption{\textbf{\model{EfficientNet} comparisons on ImageNetV2}.}
\label{tab:comparison:en:v2}\vspace{-2mm}
\end{table}

In the main paper we perform all experiments on the ImageNet~\cite{Deng2009} validation set. Here we evaluate our models on the ImageNetV2~\cite{Recht2019} test set (original test set unavailable).

\paragraph{Evaluation setup.} To study generalization of models developed on ImageNet, the authors of~\cite{Recht2019} collect a new test set following the original procedure (ImageNetV2). They find that the overall model ranks are preserved on the new test set. The absolute errors, however, increase. We repeat the comparisons from \S\ref{sec:soa} on the ImageNetV2 test set.

\paragraph{\model{ResNe(X)t} comparisons.} We compare to \model{ResNe(X)t} models in Table~\ref{tab:comparison:resnet:v2}. We observe that while model ranks are generally consistent, the gap between them decreases. Nevertheless, \model{RegNetX} models still compare favorably, and provide good models \emph{across flop regimes}, including in low-compute regimes where good \model{ResNe(X)t} models are not available. Best results can be achieved using \model{RegNetY}.

\paragraph{\model{EfficientNet} comparisons.} We compare to \model{EfficientNet} models in Table~\ref{tab:comparison:en:v2}. As before, we observe that the model ranks are generally consistent but the gap decreases. Overall, the results confirm that the \model{RegNet} models perform comparably to state-of-the-art \model{EfficientNet} while being up to \emph{5$\x$ faster} on GPUs.

\section*{Appendix B: Additional Ablations}

\begin{figure}[t]\centering
\includegraphics[height=20mm]{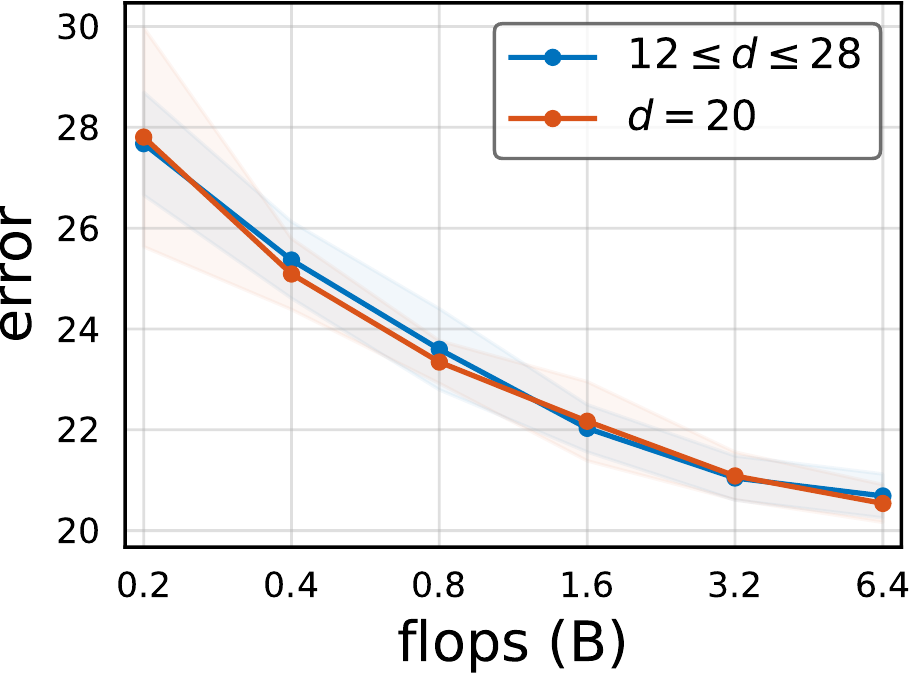}
\includegraphics[height=20mm]{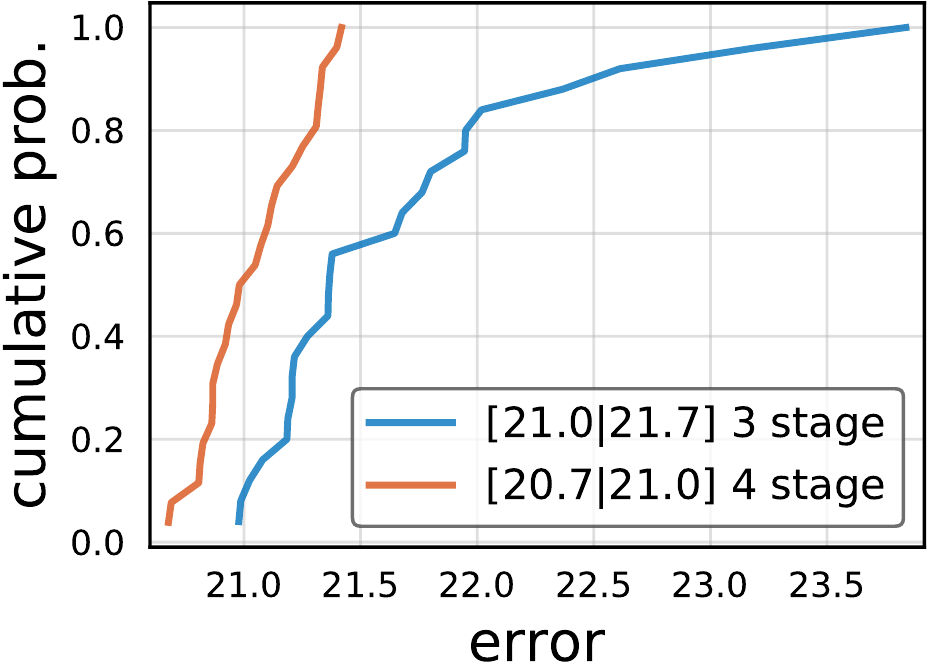}
\includegraphicstriml{40}{height=20mm}{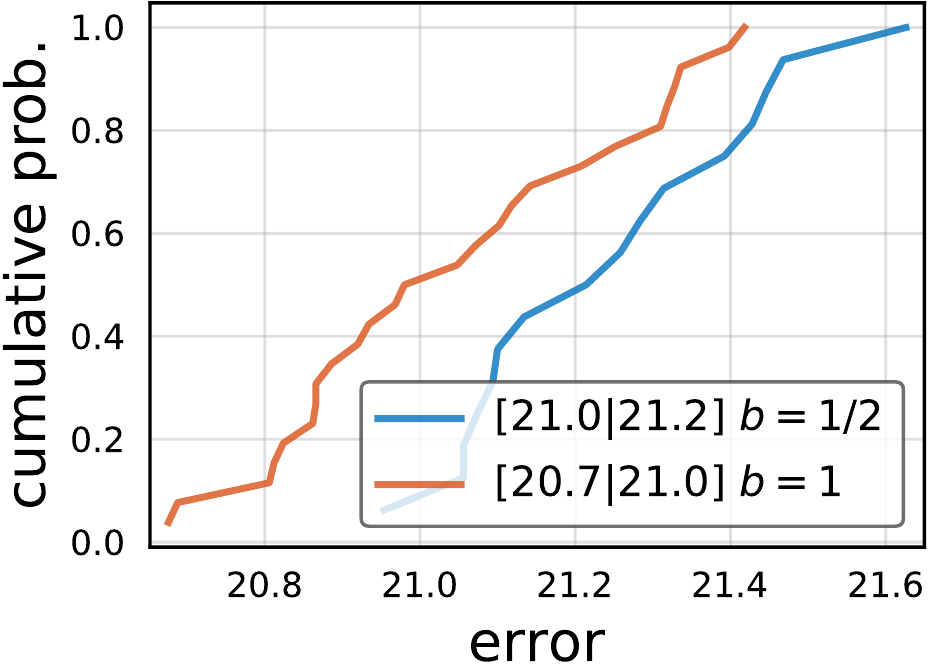}
\caption{\textbf{Additional ablations}. \textit{Left:} Fixed depth networks ($d=20$) are effective across flop regimes. \textit{Middle:} Three stage networks perform poorly at high flops. \textit{Right:} Inverted bottleneck ($b<1$) is also ineffective at high flops. See text for more context.}
\label{fig:supporting:100ep}
\end{figure}

\begin{figure}[t]\centering
\includegraphics[height=20.5mm]{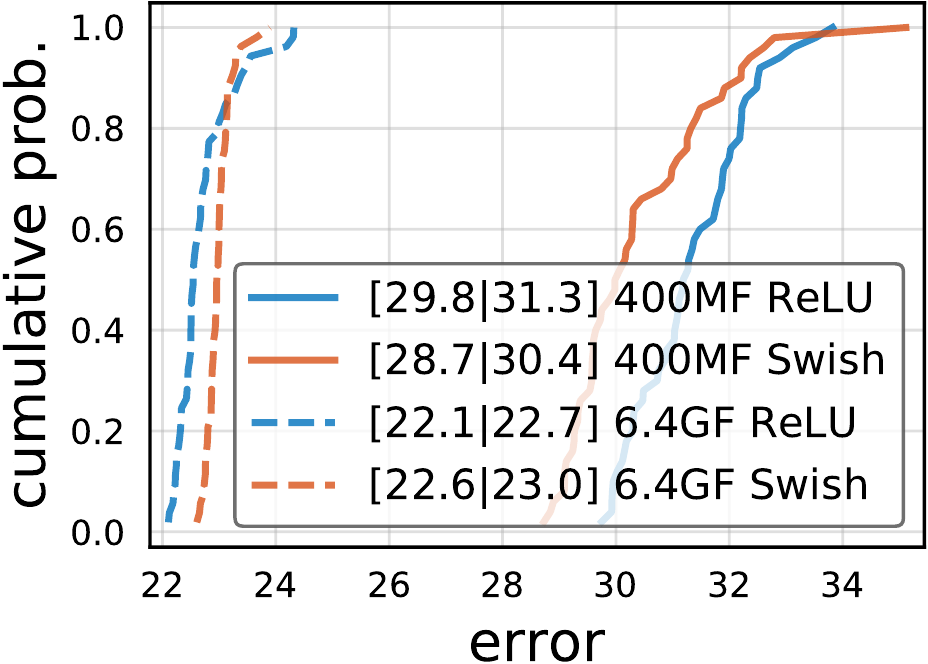}
\includegraphics[height=20.5mm]{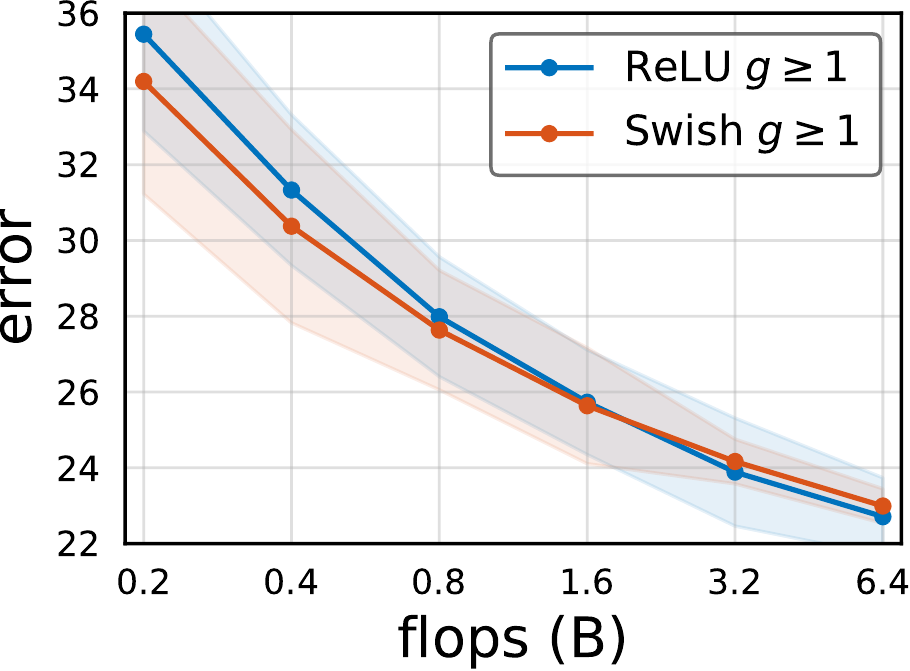}
\includegraphicstriml{32}{height=20.5mm}{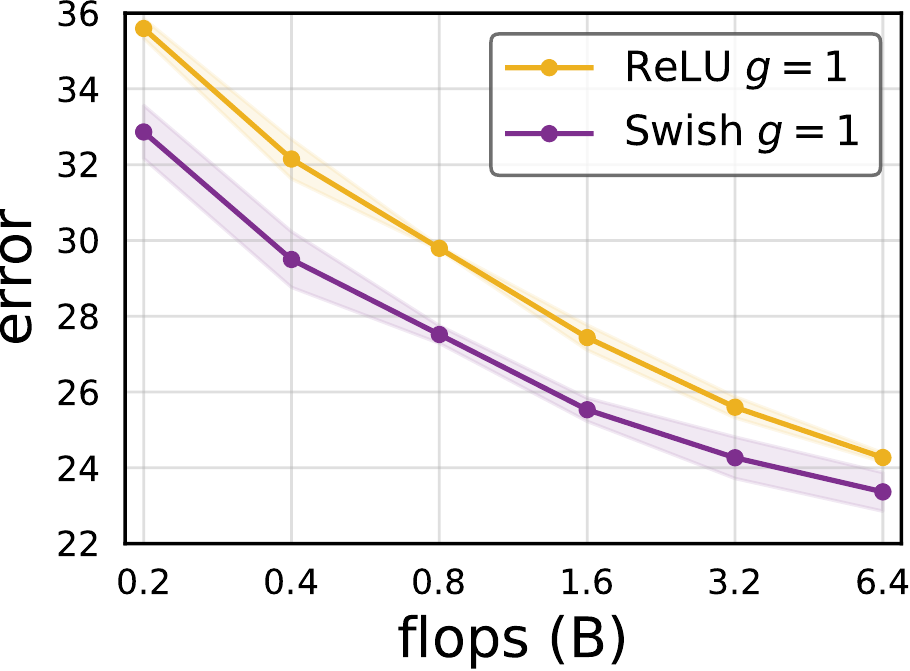}
\caption{\textbf{Swish \vs ReLU.} \textit{Left:} \regnety performs better with Swish than ReLU at 400MF but worse at 6.4GF. \textit{Middle:} Results across wider flop regimes show similar trends. \textit{Right:} If, however, $g$ is restricted to be 1 (depthwise conv), Swish is much better.}
\label{fig:supporting:swish}\vspace{-2mm}
\end{figure}

In this section we perform additional ablations to further support or supplement the results of the main text.

\paragraph{Fixed depth.} In \S\ref{sec:soa} we observed that the depths of our top models are fairly stable ($\app20$ blocks). In Figure~\ref{fig:supporting:100ep} (left) we compare using fixed depth ($d=20$) across flop regimes. To compare to our best results, we trained each model for 100 epochs. Surprisingly, we find that \emph{fixed-depth networks can match the performance of variable depth networks} for all flop regimes, in both the average and best case. Indeed, these fixed depth networks match our best results in \S\ref{sec:soa}.

\paragraph{Fewer stages.} In \S\ref{sec:soa} we observed that the top \model{RegNet} models at high flops have few blocks in the fourth stage (one or two). Hence we tested 3 stage networks at 6.4GF, trained for 100 epochs each. In Figure~\ref{fig:supporting:100ep} (middle), we show the results and observe that the three stage networks perform considerably worse. We note, however, that additional changes (\eg, in the stem or head) may be necessary for three stage networks to perform well (left for future work).

\paragraph{Inverted Bottleneck.} In \S\ref{sec:regnet} we observed that using the inverted bottleneck ($b<1$) degrades performance. Since our results were in a low-compute regime, in Figure~\ref{fig:supporting:100ep} (right) we re-test at 6.4GF and 100 epochs. Surprisingly, we find that in this regime $b<1$ degrades results further.

\paragraph{Swish \vs ReLU} Many recent methods employ the Swish~\cite{Ramachandran2017} activation function, \eg~\cite{Tan2019}. In Figure~\ref{fig:supporting:swish}, we study \regnety with Swish and ReLU. We find that Swish outperforms ReLU at low flops, but ReLU is better at high flops. Interestingly, if $g$ is restricted to be 1 (depthwise conv), Swish performs much better than ReLU. This suggests that depthwise conv and Swish interact favorably, although the underlying reason is not at all clear.

\section*{Appendix C: Optimization Settings}

\begin{figure}[t]\centering
\includegraphics[width=.326\linewidth]{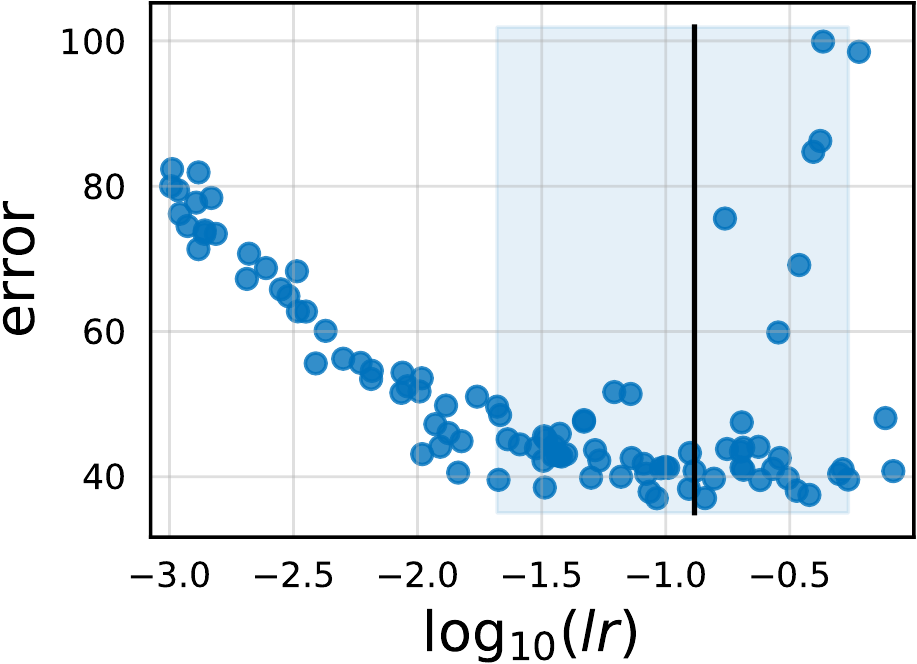}
\includegraphics[width=.326\linewidth]{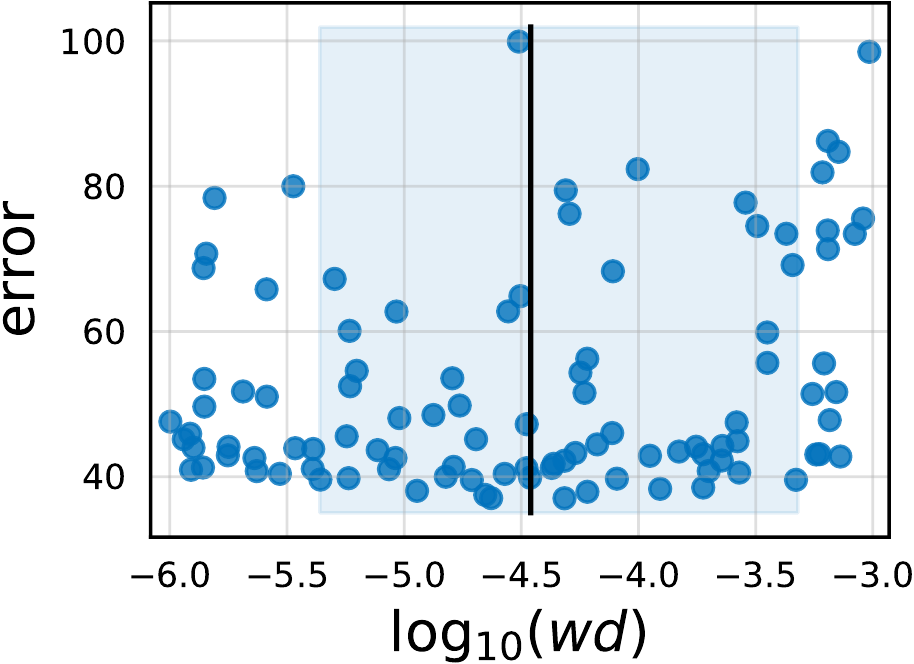}
\includegraphics[width=.326\linewidth]{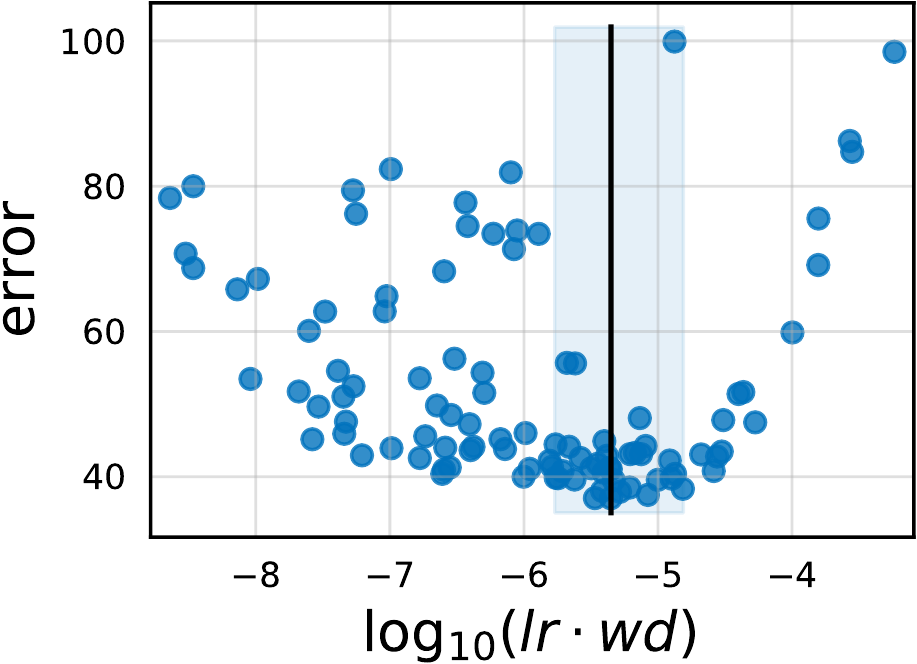}\\
\includegraphics[width=\linewidth]{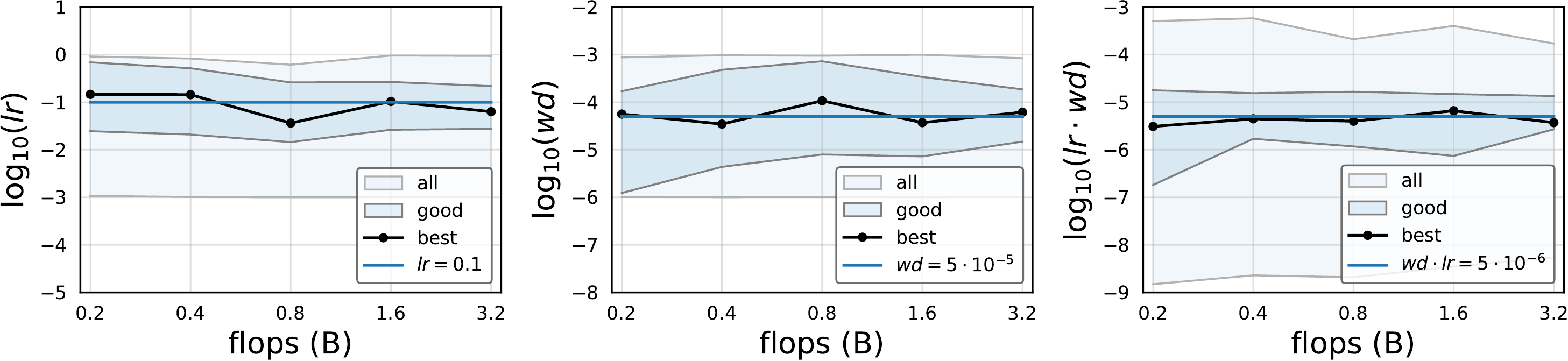}\\
\includegraphics[width=\linewidth]{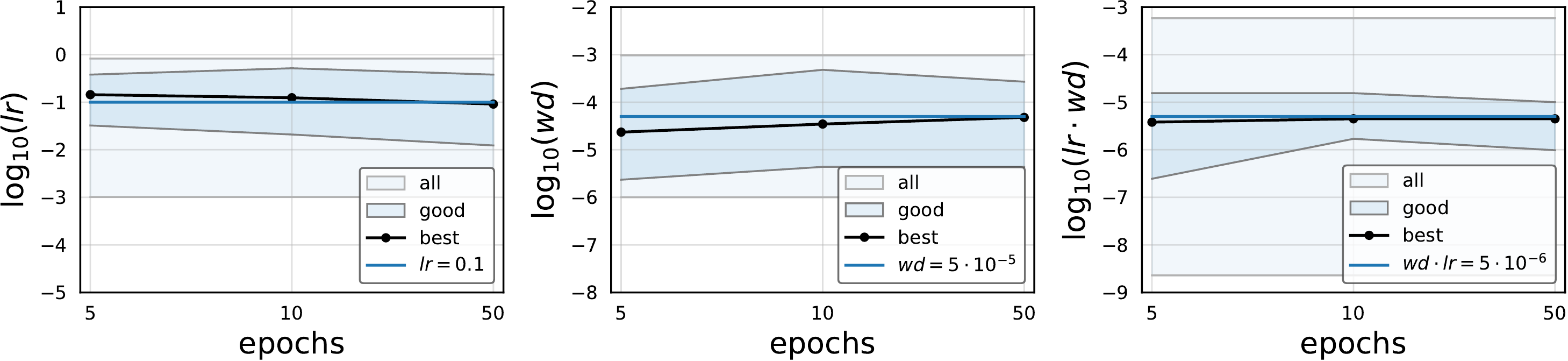}
\caption{\textbf{Optimization settings.} For these results, we generate a population of \regnetx models while also randomly varying the initial learning rate ($lr$) and weight decay ($wd$) for each model. These results use a batch size of 128 and are trained on 1 GPU. \textit{Top}: Distribution of model error versus $lr$, $wd$, and also $lr \c wd$ (at 10 epochs and 400MF). Applying an empirical bootstrap, we see that clear trends emerge, especially for $lr$ and $lr \c wd$. \textit{Middle}: We repeat this experiment but across various flop regimes (trained for 10 epochs each); the trends are stable. \textit{Bottom}: Similarly, we repeat the above while training for various number of epochs (in the 400MF regime), and observe the same trends. Based on these results, we use an $lr=0.1$ and $wd=5\c10^{-5}$ starting with \S\ref{sec:regnet} across all training schedules and flop regimes.}
\label{fig:optimization:regnet}\vspace{0mm}
\end{figure}

\begin{figure}[t]\centering
\includegraphics[width=\linewidth]{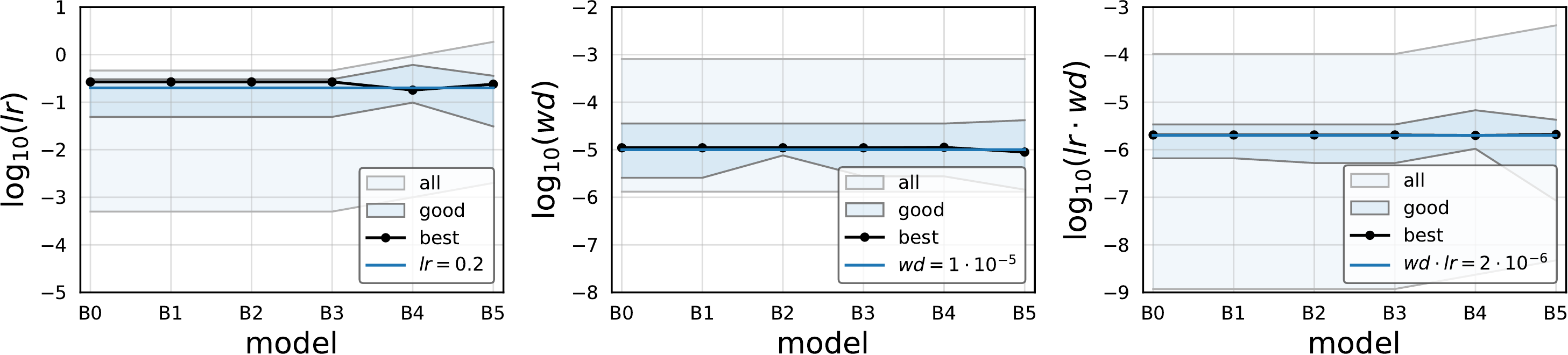}
\caption{We repeat the sweep over \textbf{$lr$ and $wd$ for \model{EfficientNet}} training each model for 25 epochs. The $lr$ (with reference to a batch size of 128) and $wd$ is stable across regimes. We use these values for all \model{EfficientNet} experiments in the main text (adjusting for batch size accordingly). See Figure~\ref{fig:optimization:regnet} for comparison.}
\label{fig:optimization:en}\vspace{-4mm}
\end{figure}

Our basic training settings follow~\cite{Radosavovic2019} as discussed in \S\ref{sec:dds}. To tune the learning rate $lr$ and weight decay $wd$ for \model{RegNet} models, we perform a study, described in Figure~\ref{fig:optimization:regnet}. Based on this, we set $lr=0.1$ and $wd=5\c10^{-5}$ for all models in \S\ref{sec:regnet} and \S\ref{sec:soa}. To enable faster training of our final models at 100 epochs, we increase the number of GPUs to 8, while keeping the number of images per GPU fixed. When scaling the batch size, we adjust $lr$ using the linear scaling rule and apply 5 epoch gradual warmup~\cite{Goyal2017}.

To enable fair comparisons, we repeat the same optimization for \model{EfficientNet} in Figure \ref{fig:optimization:en}. Interestingly, learning rate and weight decay are again stable across complexity regimes. Finally, in Table ~\ref{tab:optimization:en:b0} we report the sizable effect of training enhancement on \model{EfficientNet-B0}. The gap may be even larger for larger models (see Table~\ref{tab:comparison:en}).

\begin{table}[t]\centering
\resizebox{.95\columnwidth}{!}{\tablestyle{3pt}{1.05}
\begin{tabular}{@{}m|ccc|cc|c@{}}
 & flops (B) & params (M) & acts (M) & epochs & enhance & error \\\shline
 EfficientNet-B0                & 0.39 & 5.3 & 6.7 & 100 &             & 25.6 \\
 EfficientNet-B0                & 0.39 & 5.3 & 6.7 & 250 &             & 25.0 \\
 EfficientNet-B0                & 0.39 & 5.3 & 6.7 & 250 & \checkmark  & 24.4 \\
 EfficientNet-B0~\cite{Tan2019} & 0.39 & 5.3 & 6.7 & 350 & \checkmark\checkmark\checkmark & 23.7 \\
\end{tabular}}\vspace{1.5mm}
\caption{\textbf{Training enhancements} to \model{EfficientNet-B0}. Our \model{EfficientNet-B0} reproduction with DropPath~\cite{Larsson2017} and a 250 epoch training schedule (third row), achieves results slightly inferior to original results (bottom row), which additionally used RMSProp~\cite{Tieleman2012}, AutoAugment~\cite{Cubuk2018}, \etc. Without these enhancements to the training setup results are \app2\% lower (top row), highlighting the importance of carefully controlling the training setup.}
\label{tab:optimization:en:b0}\vspace{-1mm}
\end{table}

\begin{figure}[t]\centering
\includegraphics[width=.63\linewidth]{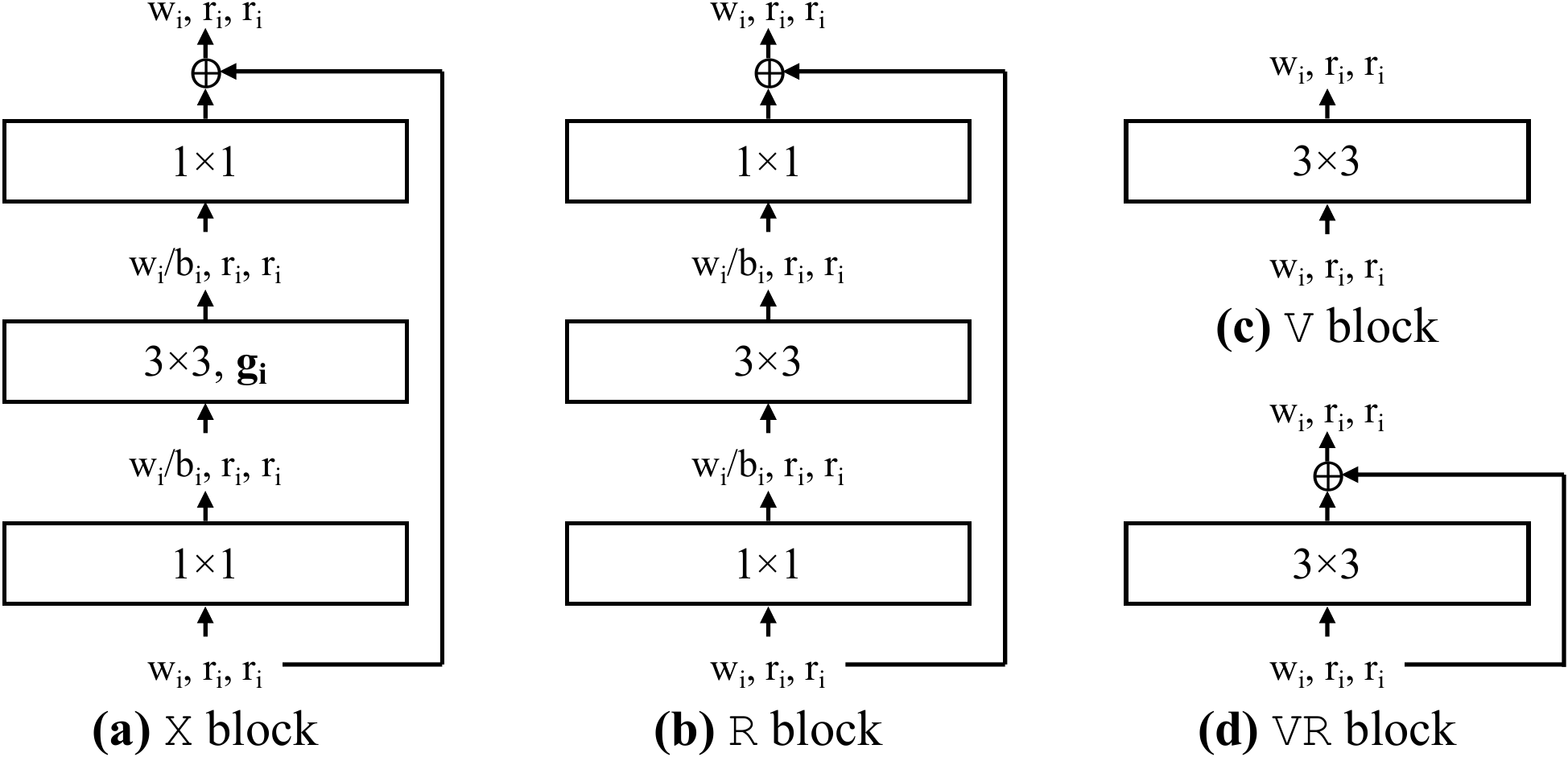}\hspace{2mm}
\includegraphics[width=.33\linewidth]{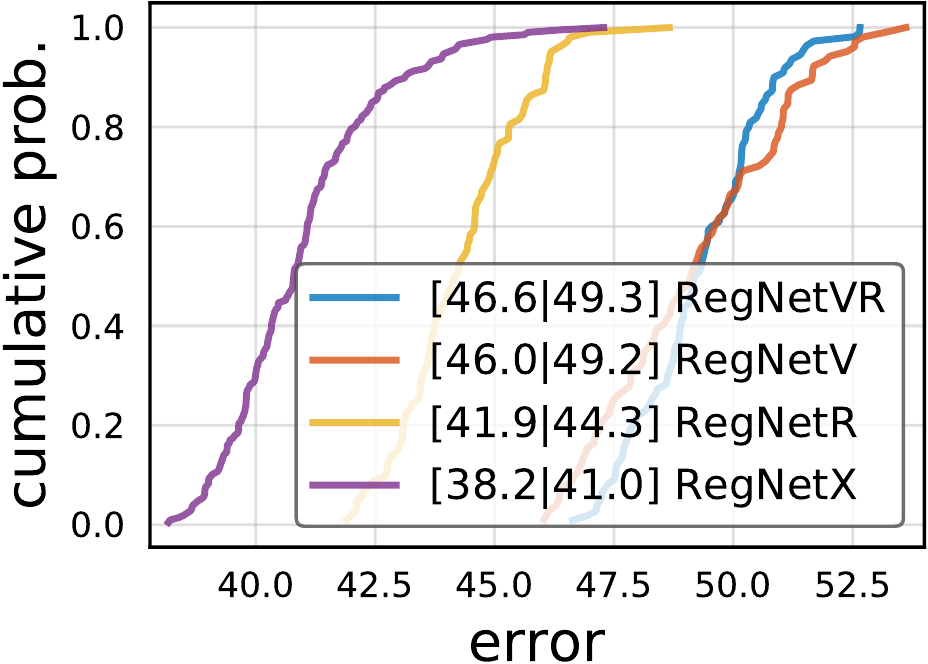}
\caption{\textbf{Block types} used in our generalization experiments (see \S\ref{sec:dds:generalize} and  Figure~\ref{fig:regnetx:generalize}). \textit{Left:} Block diagrams. \textit{Right:} Comparison of \regnet with the four block types. The \dsname{X} block performs best. Interestingly, for the \dsname{V} block, residual connections give no gain.}
\label{fig:structure:blocks}\vspace{-2mm}
\end{figure}

\section*{Appendix D: Implementation Details}

We conclude with additional implementation details.

\paragraph{Group width compatibility.} When sampling widths $w$ and groups widths $g$ for our models, we may end up with incompatible values (\ie $w$ not divisible by $g$). To address this, we employ a simple strategy. Namely, we set $g=w$ if $g > w$ and round $w$ to be divisible by $g$ otherwise. The final $w$ can be at most 1/3 different from the original $w$ (proof omitted). For models with bottlenecks, we apply this strategy to the bottleneck width instead (and adjust widths accordingly).

\paragraph{Group width ranges.} As discussed in \S\ref{sec:regnet}, we notice the general trend that the group widths of good models are larger in higher compute regimes. To account for this, we gradually adjust the group width ranges for higher compute regimes. For example, instead of sampling {\small $g \le 32$}, at 3.2GF we use {\small $16 \le g \le 64$} and allow any $g$ divisible by 8. 

\paragraph{Block types.} In \S\ref{sec:dds}, we showed that the \regnet design space generalizes to different block types. We describe these additional block types, shown in Figure~\ref{fig:structure:blocks}, next:

\begin{enumerate}[topsep=2pt, itemsep=-1ex]
\item \dsname{R} block: same as the \dsname{X} block except without groups,
\item \dsname{V} block: a basic block with only a single 3$\x$3 conv,
\item \dsname{VR} block: same as \dsname{V} block plus residual connections.
\end{enumerate}
We note that good parameter values may differ across block types. \Eg, in contrast to the \dsname{X} block, for the \dsname{R} block using $b > 1$ is better than $b = 1$. Our approach is robust to this.

\paragraph{\dsname{Y} block details.} To obtain the \dsname{Y} block, we add the SE op after the $3 \x 3$ conv of the \dsname{X} block, and we use an SE reduction ratio of $1/4$. We experimented with these choices but found that they performed comparably (not shown).

{\small\bibliographystyle{ieee}\bibliography{dds}}

\begin{thebibliography}{10}\itemsep=-1pt

\bibitem{Chollet2017}
F.~Chollet.
\newblock Xception: Deep learning with depthwise separable convolutions.
\newblock In {\em CVPR}, 2017.

\bibitem{Cubuk2018}
E.~D. Cubuk, B.~Zoph, D.~Mane, V.~Vasudevan, and Q.~V. Le.
\newblock {AutoAugment}: Learning augmentation policies from data.
\newblock {\em arXiv:1805.09501}, 2018.

\bibitem{Deng2009}
J.~Deng, W.~Dong, R.~Socher, L.-J. Li, K.~Li, and L.~Fei-Fei.
\newblock Imagenet: A large-scale hierarchical image database.
\newblock In {\em CVPR}, 2009.

\bibitem{Devries2017}
T.~DeVries and G.~W. Taylor.
\newblock Improved regularization of convolutional neural networks with cutout.
\newblock {\em arXiv:1708.04552}, 2017.

\bibitem{Efron1994}
B.~Efron and R.~J. Tibshirani.
\newblock {\em An introduction to the bootstrap}.
\newblock CRC press, 1994.

\bibitem{Goyal2017}
P.~Goyal, P.~Doll{\'a}r, R.~Girshick, P.~Noordhuis, L.~Wesolowski, A.~Kyrola,
  A.~Tulloch, Y.~Jia, and K.~He.
\newblock Accurate, large minibatch sgd: Training imagenet in 1 hour.
\newblock {\em arXiv:1706.02677}, 2017.

\bibitem{He2015b}
K.~He, X.~Zhang, S.~Ren, and J.~Sun.
\newblock Delving deep into rectifiers: Surpassing human-level performance on
  imagenet classification.
\newblock In {\em ICCV}, 2015.

\bibitem{He2016}
K.~He, X.~Zhang, S.~Ren, and J.~Sun.
\newblock Deep residual learning for image recognition.
\newblock In {\em CVPR}, 2016.

\bibitem{Howard2017}
A.~G. Howard, M.~Zhu, B.~Chen, D.~Kalenichenko, W.~Wang, T.~Weyand,
  M.~Andreetto, and H.~Adam.
\newblock Mobilenets: Efficient convolutional neural networks for mobile vision
  applications.
\newblock {\em arXiv:1704.04861}, 2017.

\bibitem{Hu2018}
J.~Hu, L.~Shen, and G.~Sun.
\newblock Squeeze-and-excitation networks.
\newblock In {\em CVPR}, 2018.

\bibitem{Huang2017}
G.~Huang, Z.~Liu, L.~Van Der~Maaten, and K.~Q. Weinberger.
\newblock Densely connected convolutional networks.
\newblock In {\em CVPR}, 2017.

\bibitem{Ioffe2015}
S.~Ioffe and C.~Szegedy.
\newblock Batch normalization: Accelerating deep network training by reducing
  internal covariate shift.
\newblock In {\em ICML}, 2015.

\bibitem{Krizhevsky2012}
A.~Krizhevsky, I.~Sutskever, and G.~E. Hinton.
\newblock Imagenet classification with deep convolutional neural networks.
\newblock In {\em NIPS}, 2012.

\bibitem{Larsson2017}
G.~Larsson, M.~Maire, and G.~Shakhnarovich.
\newblock Fractalnet: Ultra-deep neural networks without residuals.
\newblock In {\em ICLR}, 2017.

\bibitem{LeCun1989}
Y.~LeCun, B.~Boser, J.~S. Denker, D.~Henderson, R.~E. Howard, W.~Hubbard, and
  L.~D. Jackel.
\newblock Backpropagation applied to handwritten zip code recognition.
\newblock {\em Neural computation}, 1989.

\bibitem{Lee2015}
C.-Y. Lee, S.~Xie, P.~Gallagher, Z.~Zhang, and Z.~Tu.
\newblock Deeply-supervised nets.
\newblock In {\em AISTATS}, 2015.

\bibitem{Liu2018}
C.~Liu, B.~Zoph, M.~Neumann, J.~Shlens, W.~Hua, L.-J. Li, L.~Fei-Fei,
  A.~Yuille, J.~Huang, and K.~Murphy.
\newblock Progressive neural architecture search.
\newblock In {\em ECCV}, 2018.

\bibitem{Liu2019}
H.~Liu, K.~Simonyan, and Y.~Yang.
\newblock Darts: Differentiable architecture search.
\newblock In {\em ICLR}, 2019.

\bibitem{Ma2018}
N.~Ma, X.~Zhang, H.-T. Zheng, and J.~Sun.
\newblock Shufflenet v2: Practical guidelines for efficient cnn architecture
  design.
\newblock In {\em ECCV}, 2018.

\bibitem{Pham2018}
H.~Pham, M.~Y. Guan, B.~Zoph, Q.~V. Le, and J.~Dean.
\newblock Efficient neural architecture search via parameter sharing.
\newblock In {\em ICML}, 2018.

\bibitem{Radosavovic2019}
I.~Radosavovic, J.~Johnson, S.~Xie, W.-Y. Lo, and P.~Doll{\'a}r.
\newblock On network design spaces for visual recognition.
\newblock In {\em ICCV}, 2019.

\bibitem{Ramachandran2017}
P.~Ramachandran, B.~Zoph, and Q.~V. Le.
\newblock Searching for activation functions.
\newblock {\em arXiv:1710.05941}, 2017.

\bibitem{Real2018}
E.~Real, A.~Aggarwal, Y.~Huang, and Q.~V. Le.
\newblock Regularized evolution for image classifier architecture search.
\newblock In {\em AAAI}, 2019.

\bibitem{Recht2019}
B.~Recht, R.~Roelofs, L.~Schmidt, and V.~Shankar.
\newblock Do imagenet classifiers generalize to imagenet?
\newblock {\em arXiv:1902.10811}, 2019.

\bibitem{Sandler2018}
M.~Sandler, A.~Howard, M.~Zhu, A.~Zhmoginov, and L.-C. Chen.
\newblock Mobilenetv2: Inverted residuals and linear bottlenecks.
\newblock In {\em CVPR}, 2018.

\bibitem{Simonyan2015}
K.~Simonyan and A.~Zisserman.
\newblock Very deep convolutional networks for large-scale image recognition.
\newblock In {\em ICLR}, 2015.

\bibitem{Szegedy2015}
C.~Szegedy, W.~Liu, Y.~Jia, P.~Sermanet, S.~Reed, D.~Anguelov, D.~Erhan,
  V.~Vanhoucke, and A.~Rabinovich.
\newblock Going deeper with convolutions.
\newblock In {\em CVPR}, 2015.

\bibitem{Szegedy2016a}
C.~Szegedy, V.~Vanhoucke, S.~Ioffe, J.~Shlens, and Z.~Wojna.
\newblock Rethinking the inception architecture for computer vision.
\newblock In {\em CVPR}, 2016.

\bibitem{Tan2019}
M.~Tan and Q.~V. Le.
\newblock Efficientnet: Rethinking model scaling for convolutional neural
  networks.
\newblock {\em ICML}, 2019.

\bibitem{Tieleman2012}
T.~Tieleman and G.~Hinton.
\newblock Lecture 6.5-rmsprop: Divide the gradient by a running average of its
  recent magnitude.
\newblock {\em Coursera: Neural networks for machine learning}, 2012.

\bibitem{Xie2017}
S.~Xie, R.~Girshick, P.~Doll{\'a}r, Z.~Tu, and K.~He.
\newblock Aggregated residual transformations for deep neural networks.
\newblock In {\em CVPR}, 2017.

\bibitem{Zagoruyko2016}
S.~Zagoruyko and N.~Komodakis.
\newblock Wide residual networks.
\newblock In {\em BMVC}, 2016.

\bibitem{Zhang2018}
X.~Zhang, X.~Zhou, M.~Lin, and J.~Sun.
\newblock Shufflenet: An extremely efficient convolutional neural network for
  mobile devices.
\newblock In {\em CVPR}, 2018.

\bibitem{Zoph2017}
B.~Zoph and Q.~V. Le.
\newblock Neural architecture search with reinforcement learning.
\newblock In {\em ICLR}, 2017.

\bibitem{Zoph2018}
B.~Zoph, V.~Vasudevan, J.~Shlens, and Q.~V. Le.
\newblock Learning transferable architectures for scalable image recognition.
\newblock In {\em CVPR}, 2018.

\end{thebibliography}

\end{document}